\documentclass[]{TEAI}

\usepackage{helvet}
\usepackage[utf8]{inputenc}
\usepackage[T1]{fontenc}
\usepackage{array}
\usepackage{float}
\usepackage{wrapfig}
\usepackage{needspace}
\usepackage{url}
\usepackage{xurl}
\usepackage{tabularx}
\usepackage{colortbl}
\title{SCENARIO: Physically Controllable Video Generation\\with Physical Scene Guidance}

\author{
    Conghan~Yue\textsuperscript{1},
    Yuanjie~Chen\textsuperscript{1},
    Songlin~Fan\textsuperscript{1},
    Yue~Han\textsuperscript{1}, \\
    WeiYao~Zhang\textsuperscript{1},
    Chenchi~Luo\textsuperscript{2},
    Zhineng~Chen\textsuperscript{1,$\dagger$}
}

\affiliation[1]{\mbox{Institute of Trustworthy Embodied AI, Fudan University}}
\affiliation[2]{\mbox{China Mobile Shanghai ICT Co., Ltd.}}

\abstract{Physically controllable video generation requires precise control over the motion and physical interactions of objects.
Most existing methods are conditioned on motion and geometry and constrain only where objects move.
Even though other methods add physical parameters for more precise control, these parameters alone cannot determine the actual interaction, because the interaction also depends on the scene relations between an object and its surroundings.
To fill this gap, we propose \textbf{SCENARIO}, a video generation framework guided by structured physical scenes of 3D geometry, physical parameters, and object motion trajectories.
Using these trajectories as spatiotemporal anchors, it extracts local scene relations, jointly encodes them with the physical parameters, and injects the features into a pretrained video diffusion backbone.
To learn precise physical control, we build a data pipeline that strictly aligns visual assets with collision geometry and renders over 100K videos of rigid-body interactions in one-factor sweeps.
To measure this control, we establish SCENARIO-Bench and the \textit{One-Factor Physics Sweep} protocol, which compare generated and ground-truth responses to one-parameter changes.
SCENARIO outperforms prior methods on multiple benchmarks in video quality and object-region fidelity and matches the ground-truth responses most closely, demonstrating the importance of physical scene guidance for physically controllable video generation.
}

\checkdata[Code]{\url{https://github.com/Hammour-steak/SCENARIO}}
\checkdata[Website]{\url{https://scenario-video.github.io}}
\checkdata[Dataset]{\url{https://github.com/Hammour-steak/SCENARIO-Data}}
\checkdata[Bench]{\url{https://huggingface.co/datasets/Hammour/SCENARIO-Data}}

\hypersetup{
    pdftitle={SCENARIO: Physically Controllable Video Generation with Physical Scene Guidance},
    pdfauthor={Conghan Yue, Yuanjie Chen, Songlin Fan, Yue Han, WeiYao Zhang, Chenchi Luo, Zhineng Chen}
}

\begin{document}
\addtocontents{toc}{\protect\setcounter{tocdepth}{-1}}
\maketitle
\renewcommand{\thefootnote}{}
\footnotetext{$^\dagger$Corresponding author.}
\renewcommand{\thefootnote}{\arabic{footnote}}

\vspace{-1.5em}

\setcounter{topnumber}{4}
\setcounter{bottomnumber}{3}
\setcounter{totalnumber}{6}
\renewcommand{\topfraction}{0.95}
\renewcommand{\bottomfraction}{0.9}
\renewcommand{\textfraction}{0.05}
\renewcommand{\floatpagefraction}{0.95}
\setlength{\textfloatsep}{8pt plus 2pt minus 2pt}
\setlength{\intextsep}{6pt plus 1pt minus 1pt}
\setlength{\floatsep}{6pt plus 2pt minus 2pt}
\captionsetup[figure]{skip=6pt}
\captionsetup[table]{skip=6pt}
\makeatletter
\setlength{\@fptop}{0pt}
\setlength{\@fpsep}{8pt plus 1pt minus 1pt}
\setlength{\@fpbot}{0pt plus 1fil}
\makeatother
\raggedbottom

\begingroup
\makeatletter
\@clubpenalty=10000
\makeatother
\clubpenalty=10000
\section{Introduction}
\label{sec:introduction}

Video generation foundation models~\cite{vdm,svd,lumiere,cogvideox,hunyuanvideo,wan} have made significant advances in visual realism and temporal coherence, providing strong generative priors for constructing dynamic digital worlds.
However, applications such as physical world simulation, film production, and embodied AI demand physically controllable video generation, in which users require not only realistic appearance but also precise control over an object's motion within a scene and its physical interactions with surrounding objects and the environment.
In these dynamic interactions, violations of the laws of mechanics, as well as even slight geometric interpenetration or floating artifacts, significantly undermine the physical realism of generated videos.
Physically controllable video generation thus needs to generate visually natural, geometrically consistent, and physically plausible videos of scene interactions that follow the specified motion and physical conditions.

Existing studies commonly use segmentation masks~\cite{animateanything,interdyn,tavid}, depth maps~\cite{sparsectrl,videocomposer}, or motion trajectories in 2D~\cite{motionctrl,draganything} and 3D~\cite{das,3dtrajmaster,4dstreamctrl} as control signals to guide object motion by specifying spatial positions and geometric contours.
Although these geometric and motion conditions can effectively specify an object's motion path, they constrain only its position and contour over time, which the video generation model then simply follows.
In real collisions and surface contacts, an object's transitions between rolling and sliding, contact states, and collision responses are also influenced by physical parameters such as mass and the coefficients of friction and restitution.
Consequently, matching the prescribed motion trajectories or geometric changes alone cannot ensure contact and interaction details consistent with real-world physical laws.

To compensate for the limited physical information conveyed by motion and geometric conditions, recent studies have begun to condition video generation models directly on external forces or physical parameters, including localized forces and wind fields~\cite{forceprompting}, 2D physical property maps~\cite{phyco}, and object-level mass, friction, and restitution together with scene-level gravity~\cite{phyparam}.
However, physical parameters alone also cannot determine interaction outcomes, as the 3D geometry and local contact conditions of surrounding objects and the environment likewise influence the interaction process.
Identical physical parameters can produce different sliding, rebound, or multi-object collision outcomes depending on the supporting surfaces, spatial layouts, and contact configurations.
It therefore remains an important challenge to control physical parameters while keeping generated interactions consistent with the geometry and contact conditions of a given scene.

To address this challenge, we propose \textbf{SCENARIO} (physical \textbf{SCEN}e-guided controll\textbf{A}ble video gene\textbf{R}at\textbf{IO}n), a video generation framework guided by structured physical scenes of 3D geometry, physical parameters, and object motion trajectories.
It uses the motion trajectory of each object as a spatiotemporal anchor to dynamically extract the local scene relations between the object and its surroundings, jointly encodes these relations with the physical parameters, and injects the resulting features into a pretrained video diffusion backbone, providing the generator with interaction conditions that evolve with motion.
With this design, our framework combines control over motion paths with local scene constraints to guide the model in synthesizing contact and collision dynamics consistent with the geometry and physical conditions of a given scene, as \cref{fig:teaser} shows.

\begin{figure}[t]
  \centering
  \includegraphics[width=\linewidth]{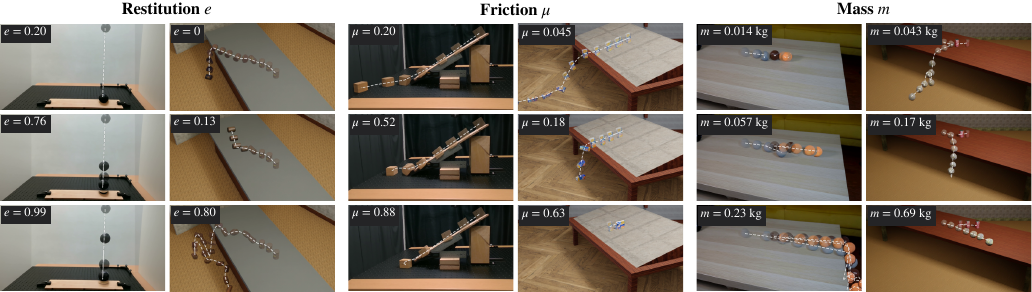}
  \caption{\textbf{Precise physical control with SCENARIO.} From a fixed first frame, the generated motion follows every value of restitution, friction, or mass from top to bottom. First frames are real in the left columns and rendered in the right columns.}
  \label{fig:teaser}
\end{figure}

To learn precise physical control, we develop a data pipeline that strictly aligns 3D visual assets with physical collision geometry and renders over 100K videos of rigid-body scenes under one-factor sweeps of their physical parameters, so that the videos of each scene differ in one physical parameter at a time.
To measure this control, we establish SCENARIO-Bench with 180 videos of 60 scenes and the \textit{One-Factor Physics Sweep} protocol, which compares how generated and ground-truth interactions change when one physical parameter changes.

Our main contributions are as follows:
(1) We propose the SCENARIO framework for video generation guided by physical scenes, using motion trajectories as anchors to jointly condition generation on dynamic local scene relations and physical parameters.
(2) We develop a data pipeline that strictly aligns 3D visual assets with physical collision geometry and renders rigid-body scenes under one-factor sweeps of their physical parameters, and we establish SCENARIO-Bench and the \textit{One-Factor Physics Sweep} protocol to assess whether generated interactions change as the ground-truth interactions do when one physical parameter changes.
(3) Experiments on multiple benchmarks and real images show that SCENARIO outperforms prior methods and follows changes in physical parameters most faithfully, demonstrating the importance of physical scene guidance for physically controllable video generation.

\endgroup
\section{Related Work}
\label{sec:related_work}

\subsection{Motion-Controllable Video Generation}

Several methods use 2D point trajectories to guide object motion~\cite{motionctrl,draganything,dragnuwa,mofavideo,motionprompting,tora,motionpro}, while Wan-Move~\cite{wanmove} propagates features of the first frame along point trajectories mapped into latent space.
AnimateAnything~\cite{animateanything}, InterDyn~\cite{interdyn}, and Kim and Joo~\cite{tavid} use masks to specify motion regions, driving entities, and interaction targets, respectively, and FlashMotion~\cite{flashmotion} adapts trajectories of masks and bounding boxes to video generation with few sampling steps.
VideoComposer~\cite{videocomposer} and SparseCtrl~\cite{sparsectrl} use depth maps and other spatial cues to control scene structure, and VACE~\cite{vace} unifies multiple visual conditioning tasks within a single framework.
These image-space signals constrain projected motion and spatial layout, but 3D object configurations and contact relations remain underdetermined.

For control beyond the image plane, DaS~\cite{das} introduces 3D tracking videos.
3DTrajMaster~\cite{3dtrajmaster} controls multiple entities through their 6DoF pose sequences, and 4DStreamCtrl~\cite{4dstreamctrl} combines 3D point trajectories, camera parameters, and depth for online control.
These approaches prescribe geometric or kinematic evolution, whereas SCENARIO adds explicit physical parameters and local scene relations to guide contact and collision dynamics.

\subsection{Physically Controllable Video Generation}

One line of work incorporates physical conditions before video synthesis through simulation or learned dynamics models~\cite{gpt4motion,physgaussian,physdreamer,physgen3d,newtongen,psivg,videophysedit}.
PhysGen~\cite{physgen} estimates physical properties and image-space geometry, simulates planar rigid-body dynamics, and refines the rendered sequence using video diffusion.
PhysCtrl~\cite{physctrl} uses physical parameters and external forces to predict 3D point trajectories, which then drive a pretrained video generator.
In both methods, physical conditions reach the video generator only through rendered motion or predicted trajectories, whereas SCENARIO also provides the physical parameters and the local scene relations along each simulated trajectory.

Other methods supply physical conditions directly to the video generator, including localized forces and wind fields in Force Prompting~\cite{forceprompting}, force goals in Goal Force~\cite{goalforce}, textual physics descriptions in PhysVid~\cite{physvid}, and 2D physical property maps in PhyCo~\cite{phyco}.
PhyParam~\cite{phyparam} uses instance masks to associate forces, mass, and the coefficients of friction and restitution with individual objects and specifies gravity at the scene level.
None of these conditions represents the 3D geometry and local contact conditions of the surroundings, which SCENARIO encodes as local scene relations along object motion trajectories.

\subsection{Physical Interaction Datasets and Benchmarks}

Simulation-based resources combine visual observations with physical state information for learning and evaluating object dynamics~\cite{videophysedit,phycove}.
CLEVRER~\cite{clevrer}, IntPhys~\cite{intphys}, and Physion~\cite{physion} support reasoning about physical events and object interactions, and ThreeDWorld~\cite{tdw} and Kubric~\cite{kubric} integrate rendering, physics simulation, and asset processing to generate richly annotated scenes.
More recent datasets introduced by PhyCo~\cite{phyco} and PhyParam~\cite{phyparam} pair simulated videos with physical parameter annotations for training controllable video generators.
Our data pipeline aligns the rendered assets with the collision geometry, records the scene geometry of every video, and renders each scene under one-factor sweeps of its physical parameters to learn precise physical control.

For evaluation, VideoPhy~\cite{videophy}, VideoPhy-2~\cite{videophy2}, PhyGenBench~\cite{phygenbench}, and PhyWorldBench~\cite{phyworldbench} assess physical commonsense in text-to-video generation.
PhyParam-Bench~\cite{phyparam} extends evaluation to explicitly parameterized rigid-body video generation, while PhysWeep~\cite{physweep} measures parameter faithfulness by recovering physical quantities from generated videos across controlled sweeps.SCENARIO-Bench measures precise physical control by comparing how generated and ground-truth interactions change when one physical parameter changes.

\section{Method}
\label{sec:method}

\begin{figure*}[t]
  \centering
  \includegraphics[width=\textwidth]{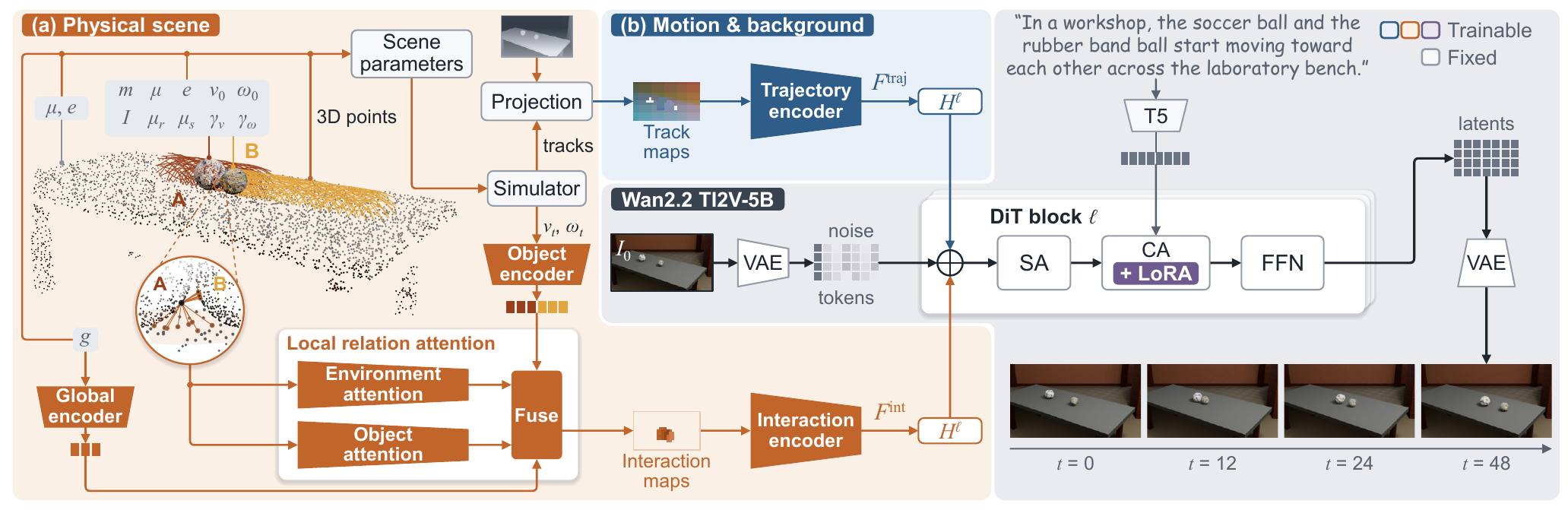}
  \caption{\textbf{Overview of SCENARIO.} (a) A rigid-body simulator turns scene parameters into 3D surface trajectories and object states. The trajectories anchor local relation attention, which encodes local scene relations with physical parameters and gravity into interaction maps and features $\mathbf{F}^{\text{int}}$. (b) Projecting the trajectories against the depth of $I_0$ yields track maps of object and background identities, encoded into $\mathbf{F}^{\text{traj}}$. Both streams enter every DiT block $\ell$ of fixed Wan2.2 TI2V-5B via zero-initialized heads $H^\ell$ with cross-attention LoRA.}
  \label{fig:framework}
\end{figure*}

Given the first frame $I_0$, a text prompt, and a structured physical scene comprising 3D scene geometry, explicit physical parameters, and the object motion trajectories simulated from them, SCENARIO synthesizes an interaction video whose contact and collision dynamics are consistent with this scene, as outlined in \cref{fig:framework}.
Each moving object is represented by persistent 3D surface points whose trajectories serve as spatiotemporal anchors to dynamically extract local scene relations with surrounding objects and the environment.
SCENARIO jointly encodes these evolving local scene relations with the physical parameters into interaction features $\mathbf{F}^{\text{int}}$ and encodes the 3D surface trajectories into trajectory features $\mathbf{F}^{\text{traj}}$, both aligned with the video tokens.
Both streams are injected into a pretrained video diffusion transformer~\cite{wan}, so that generation combines guidance on the motion path with physical constraints from the scene.

\subsection{Physical Scene and Motion Representation}
\label{sec:scene_representation}

Under a fixed camera with known intrinsics, we express all 3D quantities across frames $t \in \{0, \dots, T-1\}$ in the metric camera coordinate frame of $I_0$.
We discretize the static collision geometry into environment points $\mathbf{s}_q$ that sample the static surfaces visible in $I_0$ and extend each surface locally into the regions that the objects hide.
Every environment point carries a unit outward normal $\mathbf{n}_q$, friction $\mu_q$, and restitution $e_q$.

Each dynamic rigid object $i \in \{1, \dots, O\}$ carries mass $m_i$, sliding friction $\mu_i$, restitution $e_i$, rolling and spinning friction, linear and angular damping, an inertia tensor in its initial pose, and initial linear and angular velocities.
A rigid-body simulator integrates the motion of all objects from these parameters, the static collision geometry with its friction and restitution, and gravity $\mathbf{g}$, producing the pose, linear velocity $\mathbf{v}_{i,t}$, and angular velocity $\boldsymbol{\omega}_{i,t}$ of every object at every frame.
We sample $N_p$ points on each object surface and move them rigidly with the simulated poses, which gives persistent surface points $\{\mathbf{x}_{i,p,t}\}_{p=1}^{N_p}$ with velocities $\mathbf{v}_{i,p,t} = \mathbf{v}_{i,t} + \boldsymbol{\omega}_{i,t} \times (\mathbf{x}_{i,p,t} - \mathbf{c}_{i,t})$ about the center of mass $\mathbf{c}_{i,t}$.
Each point keeps its identity across all $T$ frames and traces a surface trajectory, and these trajectories serve as the spatiotemporal anchors.

From these quantities, an object encoder maps the mass, friction, restitution, and damping coefficients of each object, its inertia tensor rotated to the current pose, and its velocities $(\mathbf{v}_{i,t}, \boldsymbol{\omega}_{i,t})$ to object features $\mathbf{o}_{i,t}$ at every frame, and a global encoder maps $\mathbf{g}$ to a global feature $\tilde{\mathbf{g}}$.

\subsection{Trajectory-Anchored Local Scene Relations}
\label{sec:local_relations}

How an object rolls, slides, or rebounds depends on the geometry, material, and motion it encounters across its surface, so we extract these local scene relations around every surface point as the object moves.
At each frame $t$, every surface point $\mathbf{x}_{i,p,t}$ acts as an anchor that queries up to $K$ nearest neighbors within radius $r$ among the environment points to form the environment neighborhood $\mathcal{N}^{\text{env}}$, and among the surface points of other objects to form the object neighborhood $\mathcal{N}^{\text{obj}}$.
Both queries run over all points in 3D, so the hidden underside and rear points, where contact typically occurs, also find their neighbors.
Below we write $\mathbf{x}$ and $\mathbf{v}$ for the anchor and its velocity, drop the frame index from the object features, and drop all indices from the other quantities of the anchor.

Similar to the edge features of learned particle simulators~\cite{gns}, each neighbor input records the geometry, physical parameters, and relative motion that the anchor meets at one nearby point.
An environment neighbor $k \in \mathcal{N}^{\text{env}}$ is an environment point $\mathbf{s}_k$ at offset $\boldsymbol{\delta}_k = \mathbf{x} - \mathbf{s}_k$ and distance $d_k = \|\boldsymbol{\delta}_k\|$ from the anchor.
Its normal $\mathbf{n}_k$ splits the offset and the anchor velocity into normal and tangential parts, which separate the approach toward the surface from sliding along it, and the cosine $\mathbf{n}_k^\top \hat{\mathbf{g}}$ with the gravity direction $\hat{\mathbf{g}}$ tells floors, ramps, and walls apart.
Collisions instead couple the motions of two objects along their line of approach.
An object neighbor $k \in \mathcal{N}^{\text{obj}}$ is a surface point $\mathbf{x}_k$ of another object $j$ with velocity $\mathbf{v}_k$, and the offset $\boldsymbol{\delta}_k = \mathbf{x} - \mathbf{x}_k$ with length $d_k$ splits the relative velocity $\mathbf{v} - \mathbf{v}_k$ into parts along and across this line.
We write $\pi(\mathbf{u}, \mathbf{a})$ for a vector $\mathbf{u}$ with its signed component along the direction of $\mathbf{a}$ and its magnitude across that direction, and form the inputs as
\begin{equation}
\begin{aligned}
\mathbf{e}_k^{\text{env}} &= \bigl[\mathbf{n}_k, \pi(\boldsymbol{\delta}_k, \mathbf{n}_k), \pi(\mathbf{v}, \mathbf{n}_k), \mu_k, e_k, \mathbf{n}_k^\top \hat{\mathbf{g}}, \mathbf{o}_i, \tilde{\mathbf{g}}\bigr], \\
\mathbf{e}_k^{\text{obj}} &= \bigl[\boldsymbol{\delta}_k, d_k, \pi(\mathbf{v} - \mathbf{v}_k, \boldsymbol{\delta}_k), \log\tfrac{m_i}{m_j}, \mathbf{o}_i, \mathbf{o}_j, \tilde{\mathbf{g}}\bigr].
\end{aligned}
\label{eq:neighbor_inputs}
\end{equation}

Local relation attention weights each neighbor $k \in \mathcal{N}^c$ of type $c \in \{\text{env}, \text{obj}\}$ by $w_k = (1 - d_k^2 / r^2)^2$, which falls to zero at $d_k = r$.
Separate environment and object branches map each input $\mathbf{e}_k^c$ to a message $\mathbf{m}_{k,h}$ and an attention logit $a_{k,h}$ for each of $N_h$ heads and aggregate them as
\begin{equation}
\begin{aligned}
\alpha_{k,h} &= \frac{w_k \exp(a_{k,h})}{\sum_{k' \in \mathcal{N}^c} w_{k'} \exp(a_{k',h})}, \\
\mathbf{m}^c &= \Bigl(\max_{k \in \mathcal{N}^c} w_k\Bigr) \operatorname{Concat}_{h=1}^{N_h} \sum_{k \in \mathcal{N}^c} \alpha_{k,h} \mathbf{m}_{k,h}.
\end{aligned}
\label{eq:local_attention}
\end{equation}

Because the generator only sees visible surfaces, relations found at hidden contact points must reach the visible part of the same object.
For each type $c$, we apply max pooling to $\mathbf{m}^c$ over all anchors of the same object at the same frame to obtain $\bar{\mathbf{m}}^c$, map $[\mathbf{m}^c, \bar{\mathbf{m}}^c]$ through a learned projection $\Gamma^c$, and scale the result by a proximity gate $\eta^c \in [0, 1]$ shared by the whole object.
The gate takes the largest $w_k$ among the neighbors of these anchors and is zero when none of them has neighbors.
A fusion network $\Phi$ combines both gated messages with the object and global features, the anchor velocity, and $\boldsymbol{\eta} = (\eta^{\text{env}}, \eta^{\text{obj}})$ into the point interaction feature
\begin{equation}
\begin{aligned}
\tilde{\mathbf{m}}^c &= \eta^c \, \Gamma^c\bigl([\mathbf{m}^c, \bar{\mathbf{m}}^c]\bigr), \quad c \in \{\text{env}, \text{obj}\}, \\
\mathbf{f} &= \Phi\bigl(\tilde{\mathbf{m}}^{\text{env}}, \tilde{\mathbf{m}}^{\text{obj}}, \mathbf{o}_i, \tilde{\mathbf{g}}, \mathbf{v}, \boldsymbol{\eta}\bigr).
\end{aligned}
\label{eq:point_fusion}
\end{equation}
In free flight, where $\boldsymbol{\eta} = \mathbf{0}$, both gated messages vanish while $\mathbf{o}_i$, $\tilde{\mathbf{g}}$, and $\mathbf{v}$ still describe ballistic motion and spin.

\subsection{Token-Aligned Backbone Conditioning}
\label{sec:video_generation}

To bring the 3D interactions into the generated frames, SCENARIO projects all surface points onto the image plane at every frame through one shared z-buffer.
The z-buffer tests each point against the other objects and against the depth of the static scene in $I_0$, which resolves occlusions among objects and by the environment.
Each visible point falls into one cell of the spatial grid of the video latent, and the interaction maps store in each cell the mean point interaction feature $\mathbf{f}$ of the visible points of every object.

Track maps record where the same points lie in the image.
Following DaS~\cite{das}, every surface point carries a fixed identity code, its normalized image coordinates and inverse depth at frame $0$, so the code travels with the point and preserves correspondence across frames.
Each cell stores the mean code and an occupancy flag separately for every object and, in the same way, for the background pixels visible in $I_0$ that no object currently covers.

A trajectory encoder and an interaction encoder compress these maps in time and space to the latent token grid of the backbone, yielding $\mathbf{F}^{\text{traj}}$ and $\mathbf{F}^{\text{int}}$.
Because both streams share the z-buffer, the grid, and the compression, they align at every latent token.

To guide motion and contact at every depth of the transformer, we inject both streams before each of the $L$ DiT~\cite{dit} blocks $\mathcal{B}_\ell$ of the fixed Wan2.2 TI2V-5B~\cite{wan} backbone as
\begin{equation}
\mathbf{h}^{(\ell+1)} = \mathcal{B}_\ell\bigl(\mathbf{h}^{(\ell)} + H^{\ell}_{\text{traj}}(\mathbf{F}^{\text{traj}}) + H^{\ell}_{\text{int}}(\mathbf{F}^{\text{int}})\bigr),
\label{eq:block_injection}
\end{equation}
where $\mathbf{h}^{(\ell)}$ denotes the tokens entering block $\ell$, and the independent heads $H^{\ell}_{\text{traj}}$ and $H^{\ell}_{\text{int}}$ are initialized to zero so that training begins from the pretrained backbone.
We also attach LoRA~\cite{lora} adapters to the cross-attention layers.

\subsection{Training and Inference}
\label{sec:training_inference}

We train SCENARIO with rectified flow matching~\cite{flowmatching,rectifiedflow} conditioned on the first frame.
The first latent frame stays fixed to the clean VAE encoding of $I_0$, the later latent frames of the clean video latent $\mathbf{z}_0$ are mixed with standard Gaussian noise $\boldsymbol{\epsilon}$ as $\mathbf{z}_\sigma = (1 - \sigma)\mathbf{z}_0 + \sigma\boldsymbol{\epsilon}$, and the network $\hat{\mathbf{u}}_\theta$ predicts the velocity $\boldsymbol{\epsilon} - \mathbf{z}_0$.
Moving objects change both the regions they enter and the regions they vacate, while static background covers most of the frame.
We therefore split the latent positions $\tau$ of the later latent frames into a foreground region $\mathcal{M}^{\text{fg}}$, the union of the object occupancy in frame $0$ and in the current latent frame, and its complement $\mathcal{M}^{\text{bg}}$, and weight both equally in
\begin{equation}
\mathcal{L} = \frac{1}{2}\mathbb{E}_{\sigma, \boldsymbol{\epsilon}}\!\biggl[
\tfrac{1}{|\mathcal{M}^{\text{fg}}|}\!\sum_{\tau \in \mathcal{M}^{\text{fg}}}\!\|\mathbf{r}_\tau\|^2
+ \tfrac{1}{|\mathcal{M}^{\text{bg}}|}\!\sum_{\tau \in \mathcal{M}^{\text{bg}}}\!\|\mathbf{r}_\tau\|^2
\biggr],
\label{eq:flow_loss}
\end{equation}
where $\mathbf{r}_\tau = \hat{\mathbf{u}}_\theta(\mathbf{z}_\sigma, \sigma, \mathcal{C})_\tau - (\boldsymbol{\epsilon} - \mathbf{z}_0)_\tau$ is the velocity error under all conditions $\mathcal{C}$.
We jointly train all new modules and the LoRA adapters with the Wan2.2 weights fixed.

During fast impacts, large displacements between frames blur the conditions that each latent frame compresses from several video frames.
At inference, we therefore insert rigidly interpolated poses into every frame interval in which a visible surface point moves beyond a pixel threshold and recompute all conditions on this denser timeline.
We generate the video in consecutive windows that each fit one backbone pass and start from the last frame of the previous window or from $I_0$, and keep only the frames at the original timestamps.
The appendix gives all implementation details.

\section{Data Pipeline and Benchmark}
\label{sec:data}

SCENARIO targets precise physical control, in which generated interactions follow the given physical parameters as real ones do.
To learn this control, we develop a data pipeline that renders one scene under one-factor sweeps of its physical parameters and aligns every video with the physical conditions that produce it.
To measure this control, we establish SCENARIO-Bench, which compares how generated and ground-truth interactions change when one physical parameter changes.

\subsection{Data Pipeline}
\label{sec:data_pipeline}

The pipeline samples scenes with one to three dynamic objects, simulates them with PyBullet~\cite{pybullet}, and renders them with Blender~\cite{blender}.
One key design is that the rendered scene coincides with the simulated one.
Every mesh shares one metric scale with its collision shape, the collision shape covers every visible part that objects can touch, and the renderer places every mesh at its simulated pose, so the rendered contacts show neither interpenetration nor floating.
Automatic checks after every stage keep only valid scenes.
They require the simulation to follow the recorded scene and the laws of mechanics, the planned interactions to happen, and the camera to show every object clearly, and a scene that fails is replaced by a new sample of the same type.

The other key design turns each valid scene into the base scene of a group, and one-factor sweeps vary the mass, sliding friction, or restitution of one target object while geometry, initial state, appearance, and camera stay fixed.
Training on a group therefore shows the model several outcomes of one first frame that only the physical parameters and the simulated trajectories tell apart.
The dataset holds 101{,}452 videos in 7{,}804 groups, which we split by group into training, validation, and test sets.
From the same geometry and trajectories, we compute the conditions of SCENARIO as well as the point trajectories, depth maps, and bounding boxes that other control methods use.

\subsection{SCENARIO-Bench and Evaluation Protocol}
\label{sec:benchmark}

SCENARIO-Bench evaluates groups of videos that differ in one physical parameter, built from 60 base scenes of the test split, evenly divided among scenes with one, two, and three objects.
Each base scene comes with a low and a high variant of one parameter of its target object, chosen by how far the ground-truth trajectories separate, which gives 180 videos.
Every method receives the same first frame, text prompt, and seed for the three videos of a group, together with the conditions of each variant that it supports, including the simulated motion.

The \textit{One-Factor Physics Sweep} protocol scores whether the change from a base scene to each variant in the generated videos matches the change in the ground-truth videos.
We compare the center of each moving object tracked in the generated videos with its ground-truth center.
Let $\Delta\mathbf{p}(t)$ be the offset at frame $t$ from the ground-truth center of an object in the base video to its center in the video of one variant, and let $\Delta\hat{\mathbf{p}}(t)$ be the same offset between the generated videos.
With $\langle \cdot \rangle$ averaging over the future frames in which the object is visible in both ground-truth videos and then over all moving objects and both variants, each group scores
\begin{equation}
S = \max\Bigl(0,\; 1 - \frac{\sqrt{\langle \|\Delta\hat{\mathbf{p}} - \Delta\mathbf{p}\|^2 \rangle}}{\max\bigl(\sqrt{\langle \|\Delta\mathbf{p}\|^2 \rangle},\, 1\,\text{px}\bigr)}\Bigr),
\label{eq:sweep_score}
\end{equation}
where the offsets are in pixels at the evaluation resolution and the floor of one pixel exceeds the tracking error measured on ground-truth videos.
A matching response scores one, a video that stays unchanged when the parameter barely affects the motion scores close to one, and a video that ignores a clear change scores zero.
The average over all moving objects covers the motion that the target object passes to others through collisions.
A generated object that vanishes counts at least as ignoring the change, and the benchmark score averages $S$ over all groups.

\section{Experiments}
\label{sec:experiments}

\subsection{Experimental Setup}
\label{sec:setup}

We evaluate on SCENARIO-Bench and on 100 rigid-body cases each from PhyCo-Sim~\cite{phyco} and Physion++~\cite{physionpp}.
As baselines, DaS~\cite{das}, VACE~\cite{vace}, and Wan-Move~\cite{wanmove} take point trajectories, VACE also takes depth maps, FlashMotion~\cite{flashmotion} takes bounding boxes, PhyCo~\cite{phyco} takes 2D physical property maps, and PhysGen~\cite{physgen} simulates the objects in the image plane from their physical parameters and initial velocities.
We also compare with our backbone Wan2.2 TI2V-5B~\cite{wan}, which receives only the first frame and the text prompt.
We run each baseline with its released model on inputs computed from the same simulated scene as SCENARIO.
Our figures show VACE with depth maps on the benchmarks and with point trajectories on real images.

We compare the generated videos with the ground-truth videos using PSNR, SSIM~\cite{ssim}, LPIPS~\cite{lpips}, and FVD~\cite{fvd} on full frames.
The static background covers most of every frame, so we also report PSNR, SSIM, and LPIPS inside the ground-truth bounding box of each object as object-region metrics.
Contact-region metrics evaluate the same boxes within 0.25~s of each contact in the simulation.
On SCENARIO-Bench, the sweep score $S$ of \cref{eq:sweep_score} measures how closely a method reproduces the change that a physical parameter causes.
The appendix details the benchmarks, the baselines, the metrics, and the user study.

\subsection{Comparison with Prior Methods}
\label{sec:comparison}

\begin{table}[t]
  \centering
  \footnotesize
  \setlength{\tabcolsep}{3.5pt}
  \resizebox{\linewidth}{!}{%
  \begin{tabular}{@{}llccccccccccc@{}}
    \toprule
    \multirow{2}{*}[-2.5pt]{\textbf{Method}} & \multirow{2}{*}[-2.5pt]{\textbf{Control}} & \multicolumn{4}{c}{\textbf{Full frame}} & \multicolumn{3}{c}{\textbf{Object region}} & \multicolumn{3}{c}{\textbf{Contact region}} & \multirow{2}{*}[-2.5pt]{$\boldsymbol{S}$ $\boldsymbol{\uparrow}$} \\
    \cmidrule(lr){3-6} \cmidrule(lr){7-9} \cmidrule(l){10-12}
    & & \textbf{PSNR} $\boldsymbol{\uparrow}$ & \textbf{SSIM} $\boldsymbol{\uparrow}$ & \textbf{LPIPS} $\boldsymbol{\downarrow}$ & \textbf{FVD} $\boldsymbol{\downarrow}$ & \textbf{PSNR} $\boldsymbol{\uparrow}$ & \textbf{SSIM} $\boldsymbol{\uparrow}$ & \textbf{LPIPS} $\boldsymbol{\downarrow}$ & \textbf{PSNR} $\boldsymbol{\uparrow}$ & \textbf{SSIM} $\boldsymbol{\uparrow}$ & \textbf{LPIPS} $\boldsymbol{\downarrow}$ & \\
    \midrule
    Wan2.2 TI2V-5B~\cite{wan} & None & 20.39 & 0.793 & 0.236 & 1064.0 & 14.38 & 0.684 & 0.546 & 14.89 & 0.705 & 0.497 & 0.019 \\
    PhyCo~\cite{phyco} & Property maps & 26.88 & 0.904 & 0.143 & 853.9 & 15.61 & 0.728 & 0.525 & 15.38 & 0.725 & 0.518 & 0.035 \\
    PhysGen~\cite{physgen} & Parameters & 27.68 & 0.913 & 0.091 & 443.6 & 15.76 & 0.719 & 0.489 & 15.51 & 0.709 & 0.464 & 0.099 \\
    FlashMotion~\cite{flashmotion} & Boxes & 26.39 & 0.817 & 0.113 & 540.5 & 16.89 & 0.707 & 0.400 & 16.57 & 0.703 & 0.398 & 0.324 \\
    Wan-Move~\cite{wanmove} & Trajectories & 24.49 & 0.895 & 0.129 & 724.2 & 15.42 & 0.653 & 0.372 & 15.08 & 0.649 & 0.381 & 0.484 \\
    DaS~\cite{das} & Trajectories & \underline{28.39} & \underline{0.945} & \underline{0.068} & \underline{288.8} & 15.88 & 0.723 & 0.461 & 15.51 & 0.717 & 0.445 & 0.046 \\
    VACE~\cite{vace} & Trajectories & 25.52 & 0.925 & 0.130 & 403.2 & 17.48 & 0.740 & 0.340 & \underline{17.73} & 0.744 & 0.319 & 0.567 \\
    VACE~\cite{vace} & Depth maps & 24.16 & 0.919 & 0.092 & 297.5 & \underline{17.55} & \underline{0.760} & \underline{0.288} & 17.59 & \underline{0.760} & \underline{0.287} & \underline{0.678} \\
    \midrule
    SCENARIO & Physical scene & \textbf{34.58} & \textbf{0.950} & \textbf{0.035} & \textbf{76.1} & \textbf{21.32} & \textbf{0.797} & \textbf{0.189} & \textbf{21.28} & \textbf{0.802} & \textbf{0.181} & \textbf{0.851} \\
    \bottomrule
  \end{tabular}}
  \caption{\textbf{Quantitative comparison on SCENARIO-Bench.} Control is the condition that each method receives besides the first frame and the text prompt, and $S$ is the sweep score of \cref{eq:sweep_score}. Bold marks the best result and underline the best baseline.}
  \label{tab:bench}
\end{table}

\begin{figure}[t]
  \centering
  \includegraphics[width=\textwidth]{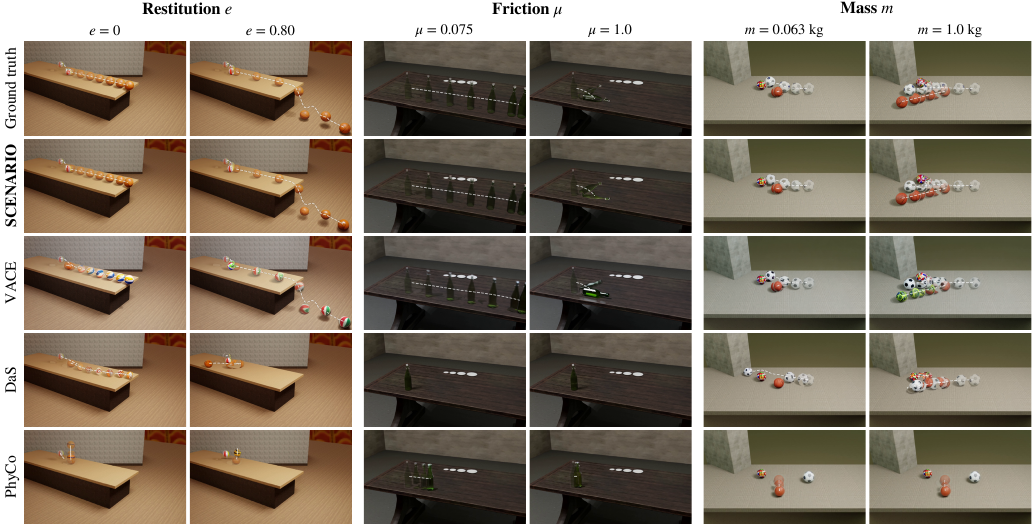}
  \caption{\textbf{Qualitative comparison on SCENARIO-Bench.} Each panel overlays one video at the same instants, and dashed lines trace the tracked objects. From left to right, the thrown beach ball knocks the basketball off the counter only with high restitution, the pushed bottle tips over only with high friction, and only the heavy soccer ball pushes the basketball away.}
  \label{fig:qualitative}
\end{figure}

\begin{table}[t]
  \centering
  \footnotesize
  \setlength{\tabcolsep}{3.5pt}
  \resizebox{\linewidth}{!}{%
  \begin{tabular}{@{}llcccccccccc@{}}
    \toprule
    \multirow{2}{*}[-2.5pt]{\textbf{Benchmark}} & \multirow{2}{*}[-2.5pt]{\textbf{Method}} & \multicolumn{4}{c}{\textbf{Full frame}} & \multicolumn{3}{c}{\textbf{Object region}} & \multicolumn{3}{c}{\textbf{Contact region}} \\
    \cmidrule(lr){3-6} \cmidrule(lr){7-9} \cmidrule(l){10-12}
    & & \textbf{PSNR} $\boldsymbol{\uparrow}$ & \textbf{SSIM} $\boldsymbol{\uparrow}$ & \textbf{LPIPS} $\boldsymbol{\downarrow}$ & \textbf{FVD} $\boldsymbol{\downarrow}$ & \textbf{PSNR} $\boldsymbol{\uparrow}$ & \textbf{SSIM} $\boldsymbol{\uparrow}$ & \textbf{LPIPS} $\boldsymbol{\downarrow}$ & \textbf{PSNR} $\boldsymbol{\uparrow}$ & \textbf{SSIM} $\boldsymbol{\uparrow}$ & \textbf{LPIPS} $\boldsymbol{\downarrow}$ \\
    \midrule
    \multirow{5}{*}{PhyCo-Sim} & Wan2.2 TI2V-5B~\cite{wan} & 21.90 & 0.820 & 0.243 & 1055.8 & 16.13 & 0.775 & 0.496 & 15.70 & 0.806 & 0.474 \\
    & PhyCo~\cite{phyco} & 26.12 & 0.869 & 0.232 & 523.6 & 18.44 & 0.803 & 0.422 & 17.99 & 0.823 & 0.417 \\
    & VACE~\cite{vace} (depth) & 25.10 & 0.847 & 0.161 & 259.9 & \underline{22.16} & \underline{0.845} & \underline{0.193} & \underline{23.06} & \underline{0.896} & \underline{0.144} \\
    & DaS~\cite{das} & \underline{28.56} & \underline{0.936} & \underline{0.095} & \underline{210.9} & 19.30 & 0.821 & 0.340 & 19.33 & 0.857 & 0.303 \\
    \cmidrule(l){2-12}
    & SCENARIO & \textbf{33.52} & \textbf{0.939} & \textbf{0.064} & \textbf{122.5} & \textbf{25.39} & \textbf{0.881} & \textbf{0.126} & \textbf{24.25} & \textbf{0.906} & \textbf{0.125} \\
    \midrule
    \multirow{5}{*}{Physion++} & Wan2.2 TI2V-5B~\cite{wan} & 16.62 & 0.509 & 0.466 & 1527.5 & 13.90 & 0.607 & 0.651 & 13.93 & 0.612 & 0.645 \\
    & PhyCo~\cite{phyco} & 23.99 & 0.773 & 0.249 & 579.3 & 15.42 & 0.690 & 0.557 & 15.39 & 0.689 & 0.557 \\
    & VACE~\cite{vace} (depth) & 22.79 & 0.757 & 0.174 & 514.4 & \underline{17.50} & \underline{0.691} & \underline{0.374} & \underline{17.33} & 0.688 & \underline{0.383} \\
    & DaS~\cite{das} & \underline{27.66} & \underline{\textbf{0.903}} & \underline{0.114} & \underline{403.9} & 16.25 & \underline{0.727} & 0.477 & 16.38 & \underline{0.728} & 0.471 \\
    \cmidrule(l){2-12}
    & SCENARIO & \textbf{29.10} & 0.854 & \textbf{0.066} & \textbf{278.6} & \textbf{19.95} & \textbf{0.747} & \textbf{0.309} & \textbf{20.00} & \textbf{0.750} & \textbf{0.307} \\
    \bottomrule
  \end{tabular}}
  \caption{\textbf{Quantitative comparison on PhyCo-Sim and Physion++.} Bold marks the best result and underline the best baseline.}
  \label{tab:external}
\end{table}

\Cref{tab:bench} shows that only SCENARIO keeps full-frame quality while following changes in the physical parameter.
Compared with the strongest baseline on each metric, SCENARIO lowers FVD to about a quarter and raises the object-region PSNR by 3.8~dB.
The baselines that receive the simulated motion either keep the appearance of the first frame but barely respond to the parameter, or follow the motion at the cost of full-frame quality.
The baselines without simulated motion barely respond, including those that take 2D property maps or simulate the objects in the image plane.
SCENARIO encodes the physical parameters and the local scene relations together with the simulated motion, and \cref{sec:ablation} shows that each of them improves every metric.
In \cref{fig:qualitative}, SCENARIO reproduces the outcome of each value and keeps the objects intact, whereas the baselines either miss the outcomes or alter the objects.

\begin{figure}[t]
  \centering
  \includegraphics[width=0.72\linewidth]{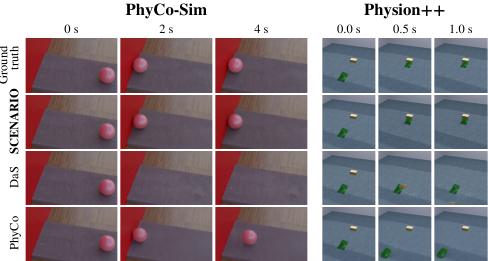}
  \caption{\textbf{Qualitative comparison on PhyCo-Sim and Physion++.} On PhyCo-Sim a ball rolls into a wall, and on Physion++ a green block slides into a yellow block.}
  \label{fig:qualitative_other}
\end{figure}

\Cref{tab:external} and \cref{fig:qualitative_other} show that SCENARIO transfers to rigid-body scenes from other datasets, and it surpasses PhyCo even on PhyCo-Sim, the dataset that PhyCo introduced.

\subsection{Results on Real Images}
\label{sec:real}

\begin{figure*}[t]
  \centering
  \includegraphics[width=\textwidth]{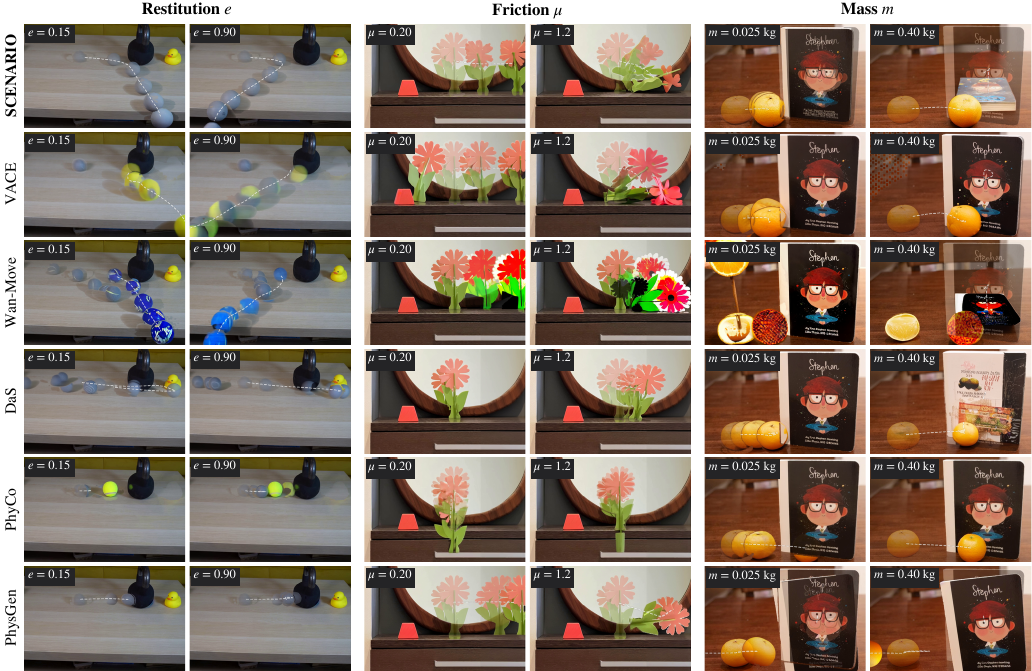}
  \caption{\textbf{Qualitative comparison on real images.} From left to right, a gray ball rolls into a kettlebell, a flower cutout is pushed along a dresser, and an orange rolls into a standing book, with low and high restitution, friction, and mass respectively.}
  \label{fig:real}
\end{figure*}

SCENARIO also generates physically controlled videos from real images.
We collect 18 scenes from Physics-IQ~\cite{physicsiq}, Physics 101~\cite{phys101}, PhysGen~\cite{physgen}, PhysGen3D~\cite{physgen3d}, and GAUGE~\cite{gauge} and build 36 one-factor sweeps of restitution, friction, and mass on them.
For each image, MoGe-2~\cite{moge2} predicts metric depth and SAM~2~\cite{sam2} segments the objects, which together give the collision geometry, and PyBullet~\cite{pybullet} simulates every sweep.
SCENARIO generates these videos with the model of \cref{sec:comparison} without any further training, although we train that model only on our rendered data.
The appendix details the scene construction and shows six more sweeps.

In \cref{fig:real}, SCENARIO again keeps the objects and changes their motion with the parameter value in every sweep, whereas each baseline misses the change or alters the objects in at least one sweep.
In a user study on all 36 sweeps, participants prefer SCENARIO over each compared baseline in at least 91.7\% of the judgments, both on which motion better fits the physical setting stated for each pair of videos and on visual quality.
From the videos of SCENARIO, they tell which variant has the heavier, bouncier, or more slippery object in 94.4\% of the judgments, against at most 77.8\% for these baselines.

\subsection{Ablation Study}
\label{sec:ablation}

\begin{table}[t]
  \centering
  \small
  \setlength{\tabcolsep}{7pt}%
  \begin{tabular}{@{}ccccccc@{}}
    \toprule
    \textbf{Physical} & \textbf{Local scene} & \multicolumn{2}{c}{\textbf{Full frame}} & \multicolumn{2}{c}{\textbf{Object region}} & \multirow{2}{*}[-2.5pt]{$\boldsymbol{S}$ $\boldsymbol{\uparrow}$} \\
    \cmidrule(lr){3-4} \cmidrule(lr){5-6}
    \textbf{parameters} & \textbf{relations} & \textbf{PSNR} $\boldsymbol{\uparrow}$ & \textbf{FVD} $\boldsymbol{\downarrow}$ & \textbf{PSNR} $\boldsymbol{\uparrow}$ & \textbf{LPIPS} $\boldsymbol{\downarrow}$ & \\
    \midrule
    $\times$ & $\times$ & 33.07 & 109.3 & 19.21 & 0.238 & 0.763 \\
    $\times$ & \checkmark & 33.16 & 101.5 & 19.58 & 0.236 & 0.785 \\
    \checkmark & $\times$ & 33.47 & 90.3 & 19.76 & 0.229 & 0.802 \\
    \checkmark & \checkmark & \textbf{34.58} & \textbf{76.1} & \textbf{21.32} & \textbf{0.189} & \textbf{0.851} \\
    \bottomrule
  \end{tabular}
  \caption{\textbf{Ablation on SCENARIO-Bench.} All models share the same training data and budget. Bold marks the best result.}
  \label{tab:ablation}
\end{table}

\Cref{tab:ablation} adds the physical parameters and the local scene relations to a model whose backbone is conditioned on the trajectory features $\mathbf{F}^{\text{traj}}$ alone, without the interaction features $\mathbf{F}^{\text{int}}$.
Each of them alone improves every metric, and the gain from adding both exceeds the sum of the two individual gains on every metric.
The local scene relations describe where an object meets its surroundings and the physical parameters determine how this contact resolves, so encoding them jointly tells the generator both where each contact occurs and how it resolves.

\section{Conclusion}
\label{sec:conclusion}

We presented SCENARIO, which conditions a pretrained video diffusion backbone on physical parameters and on local scene relations extracted along object motion trajectories, together with SCENARIO-Bench and a data pipeline that renders aligned rigid-body interactions in one-factor sweeps.
SCENARIO outperforms prior methods on multiple benchmarks in video quality and object-region fidelity, and it matches the ground-truth responses to one-parameter changes most closely.
Trained only on our rendered data, it also generates physically controlled videos from single real images.
These results show that local scene relations and physical parameters together make generated interactions more realistic, and that physical scene guidance makes physically controllable video generation more precise.

\FloatBarrier
\bibliographystyle{plainnat}
\bibliography{references}

\clearpage
\appendix
\setcounter{equation}{0}
\setcounter{figure}{0}
\setcounter{table}{0}
\renewcommand{\theequation}{S\arabic{equation}}
\renewcommand{\thefigure}{S\arabic{figure}}
\renewcommand{\thetable}{S\arabic{table}}
\renewcommand{\theHequation}{S\arabic{equation}}
\renewcommand{\theHfigure}{S\arabic{figure}}
\renewcommand{\theHtable}{S\arabic{table}}
\addtocontents{toc}{\protect\setcounter{tocdepth}{2}}
\begingroup
\renewcommand{\contentsname}{Table of Contents for Appendix}
\setstretch{1}
\fontsize{10.5}{14}\selectfont
\renewcommand{\cfttoctitlefont}{\fontsize{16}{20}\selectfont\bfseries\color{seedblue}}
\renewcommand{\cftsecfont}{\fontsize{11.5}{14.5}\selectfont\bfseries}
\renewcommand{\cftsecpagefont}{\fontsize{11.5}{14.5}\selectfont\bfseries}
\renewcommand{\cftsubsecfont}{\fontsize{10.5}{13.5}\selectfont}
\renewcommand{\cftsubsecpagefont}{\fontsize{10.5}{13.5}\selectfont}
\setlength{\cftbeforetoctitleskip}{0pt}
\setlength{\cftaftertoctitleskip}{12pt}
\setlength{\cftbeforesecskip}{7pt}
\setlength{\cftbeforesubsecskip}{2pt}
\setlength{\cftparskip}{0pt}
\cftsetindents{section}{0em}{1.8em}
\cftsetindents{subsection}{1.8em}{2.7em}
\cftsetpnumwidth{2em}
\cftsetrmarg{3em}
\tableofcontents
\endgroup
\clearpage
\raggedbottom
\section{Method Details}
\label{supp:method}

This section gives the design details of \cref{sec:method} under the subsection titles of the main text, and \cref{tab:supp_hyper} summarizes the settings.
All values are those of the complete model evaluated in \cref{sec:experiments}.

\subsection{Physical Scene and Motion Representation}
\label{supp:representation}

\paragraph{Environment points.}
We rasterize the static collision triangles from the camera of $I_0$ at pixel centers and keep the first hit of every pixel.
A first hit counts as visible when its depth agrees with the depth of the first frame within $\max(0.02\,\text{m}, 0.01z)$ at camera depth $z$ and it lies outside the object mask of the first frame.
Inside the object mask, a hidden hit joins the set only if a region grown from a visible static pixel reaches it within 16 pixels at an image height of 720 pixels and it lies on the plane of that pixel, so the hidden underside of each object rests on collision triangles that continue a visible supporting plane.
We sample the environment points uniformly over the visible pixels, reserve up to eight pixels for every thin static structure, and give each point the normal, friction, and restitution of its triangle.
During training, half of the samples drop a random fraction of up to 10\% of their environment points as augmentation.

\paragraph{Object and global encoders.}
The object encoder is an MLP with widths 19, 512, 512, and 256 and SiLU activations.
Its 19 inputs per object and frame are the mass, friction, and restitution, the six entries of the inertia tensor $\mathbf{I}_{i,t} = \mathbf{R}_{i,t} \mathbf{I}_{i,0} \mathbf{R}_{i,t}^\top$ obtained by rotating the inertia tensor $\mathbf{I}_{i,0}$ of the initial pose with the rotation $\mathbf{R}_{i,t}$ from that pose, the rolling and spinning friction, the linear and angular damping, and the current velocities $\mathbf{v}_{i,t}$ and $\boldsymbol{\omega}_{i,t}$.
We feed $\log m_i$, $\log(1+\mu_i)$, $2e_i - 1$, $\operatorname{sign}(I)\log(1 + 1000|I|)$ for every inertia entry $I$, $\log(1+x)$ for the rolling and spinning friction and the damping coefficients, $\operatorname{asinh}(\mathbf{v}_{i,t})$, and $\operatorname{asinh}(\boldsymbol{\omega}_{i,t}/5)$.
The global encoder maps $\mathbf{g}/9.81$ through an MLP with widths 3, 64, and 64.

\begin{table}[t]
  \centering
  \small
  \begin{tabular*}{0.68\linewidth}{@{\extracolsep{\fill}}ll@{}}
    \toprule
    \textbf{Setting} & \textbf{Value} \\
    \midrule
    Video & $832 \times 480$, 97 frames, 24 fps \\
    Objects per scene & 1 to 3 \\
    Environment points & 8192 \\
    Surface points per object $N_p$ & 2048 \\
    Neighbors $K$, radius $r$ & 8, 0.15\,m \\
    Attention heads $N_h$, width & 4, 512 \\
    Object, global, point features & 256, 64, 256 \\
    $\mathbf{F}^{\text{traj}}$, $\mathbf{F}^{\text{int}}$ channels & 32, 256 \\
    Head $H^{\ell}_{\text{traj}}$ & $32 \to 16 \to 3072$ \\
    Head $H^{\ell}_{\text{int}}$ & $256 \to 128 \to 3072$ \\
    DiT blocks $L$, LoRA rank & 30, 32 \\
    Sampling steps, guidance, shift & 30, 3, 5 \\
    Subdivision threshold & 16 pixels \\
    \bottomrule
  \end{tabular*}
  \caption{\textbf{Settings of SCENARIO.}}
  \label{tab:supp_hyper}
\end{table}

\subsection{Trajectory-Anchored Local Scene Relations}
\label{supp:relations}

\paragraph{Neighbor inputs.}
The input $\mathbf{e}_k^{\text{env}}$ of an environment neighbor in \cref{eq:neighbor_inputs} starts with 16 relational entries.
They are the offset $\boldsymbol{\delta}_k / r$, the normal $\mathbf{n}_k$, the signed normal offset $\boldsymbol{\delta}_k^\top \mathbf{n}_k / r$ and the tangential offset $\|\boldsymbol{\delta}_k - (\boldsymbol{\delta}_k^\top \mathbf{n}_k)\mathbf{n}_k\| / r$, $\log(1+\mu_k)$ and $e_k$, the anchor velocity $\operatorname{asinh}(\mathbf{v})$, $\operatorname{asinh}$ of its signed normal speed $\mathbf{v}^\top \mathbf{n}_k$ and its tangential speed, and the cosine $\mathbf{n}_k^\top \hat{\mathbf{g}}$.
The input $\mathbf{e}_k^{\text{obj}}$ of an object neighbor starts with 10 relational entries. They are the offset $\boldsymbol{\delta}_k / r$, the distance $d_k / r$, the relative velocity $\operatorname{asinh}(\mathbf{v} - \mathbf{v}_k)$, $\operatorname{asinh}$ of its signed speed along $\boldsymbol{\delta}_k / d_k$ and its speed across this direction, and the log mass ratio.
Both inputs then append the object feature of the anchor and the global feature, which have 256 and 64 dimensions, and the input of an object neighbor also appends the object feature of the neighbor, which has 256 dimensions.

\paragraph{Local relation attention.}
Both relation types use $K = 8$ neighbors within $r = 0.15$\,m.
Each branch maps an input $\mathbf{e}_k^c$ through three linear layers of width 512 with SiLU, SiLU, and ReLU activations to a message of 512 channels split into $N_h = 4$ heads of 128 channels, and a linear map computes the four logits from that message.
The weights $w_k$ act inside the normalized attention of \cref{eq:local_attention}, so a neighbor near the radius fades out while closer neighbors remain.
Because the normalized weights always sum to one, the attention alone would keep the message at full strength until the last neighbor leaves.
The outer factor $\max_{k \in \mathcal{N}^c} w_k$ instead scales the message to zero as even the nearest neighbor reaches the radius, which joins the empty case $\mathbf{m}^c = \mathbf{0}$ continuously.
The projection $\Gamma^c$ of \cref{eq:point_fusion} maps the 1024 channels of $[\mathbf{m}^c, \bar{\mathbf{m}}^c]$ to 256.
The fusion network $\Phi$ takes the two gated messages, $\mathbf{o}_i$, $\tilde{\mathbf{g}}$, $\operatorname{asinh}(\mathbf{v})$, and $\boldsymbol{\eta}$, 837 channels in total, through a linear layer of width 512, SiLU, and a linear layer to 256 channels, followed by a residual block with layer normalization and an MLP of two linear layers with hidden width 512.

\subsection{Token-Aligned Backbone Conditioning}
\label{supp:conditioning}

\paragraph{Track and interaction maps.}
Both maps use the spatial grid of the video latent, 52 cells wide and 30 cells high for an $832 \times 480$ video.
The projection splats every point with a radius of one pixel at $832 \times 480$, resolves overlaps among objects with the z-buffer, and keeps a point only where it lies in front of the background depth of the first frame, with the background winning ties.
A track map has 16 channels per frame, four for each of up to three objects and four for the background, and each group holds an identity code of three channels and an occupancy flag.
The identity code of a point is $(u/W, v/H, \rho)$ at frame 0, where $\rho$ normalizes the inverse depth by the 2nd and 98th percentiles of the inverse depth in the first frame.
The background channels use $(x/W, y/H, \rho)$ of every pixel that has valid depth in the first frame and lies outside the object mask of that frame, and they leave pixels covered by objects in the current frame or in $I_0$ empty.
Each cell averages all splats that fall into it.
An interaction map has $3 \times 256 = 768$ channels per frame, and each cell averages the point interaction features of the visible points of each object.

\paragraph{Encoders and injection heads.}
The trajectory encoder keeps frame 0 and compresses each later group of four frames with a temporal convolution of kernel and stride 4, initialized to the group average and trained.
The interaction encoder keeps frame 0 and averages each later group of four frames.
A $1 \times 2 \times 2$ patch convolution with SiLU then maps the 16 track channels to 32 and the 768 interaction channels to 256, on a token grid of 25 latent frames, 15 rows, and 26 columns for 97 frames.
Every one of the $L = 30$ blocks in \cref{eq:block_injection} has its own heads $H^{\ell}_{\text{traj}}$ with widths 32, 16, and 3072 and $H^{\ell}_{\text{int}}$ with widths 256, 128, and 3072, each the product of two linear maps whose output factor starts at zero.
LoRA~\cite{lora} adapters of rank 32 and alpha 32 act on the query, key, value, and output projections of the cross-attention in all 30 blocks.

\subsection{Training and Inference}
\label{supp:training}

\paragraph{Training.}
We train SCENARIO on the training split of \cref{sec:data_pipeline} for 9{,}216 steps with 8 videos per step on 4 GPUs.
We sample the noise level $\sigma$ from a shifted uniform schedule with shift 5 and keep the first latent frame clean.
The occupancy that defines $\mathcal{M}^{\text{fg}}$ comes from the object occupancy flags of the track maps, and the occupancy of a later latent frame is the union over its four video frames.
The two regions each receive half of the loss weight in \cref{eq:flow_loss}.
We use AdamW~\cite{adamw} with learning rates of $10^{-4}$ for the encoders and heads and $5 \times 10^{-5}$ for LoRA, weight decay 0.01 on the encoders, 229 warmup steps followed by a cosine schedule, gradient clipping at 1.0, and bfloat16 autocast.

\paragraph{Subdivision and windowed inference.}
For every native frame interval, we measure the largest step of a visible surface point in the $832 \times 480$ frame and insert $\lceil s / 16 \rceil - 1$ samples for a step of $s$ pixels.
The inserted samples use rigidly interpolated poses, with spherical interpolation of the rotation and linear interpolation of the center of mass, and we add samples until no step exceeds 16 pixels.
Intervals below the threshold keep a single step, so every original timestamp falls on an integer position of the subdivided timeline.
We pack consecutive intervals into windows of at most 96 steps, which gives at most 97 frames, and pad each window to $4k+1$ frames by repeating its last condition frame.
For every window, we recompute the surface points, their velocities, the object states, the local scene relations, and both kinds of maps on its timeline.
The first window starts from $I_0$, and each later window starts from the last frame that the previous window generated at an original timestamp.
Each window recomputes the identity codes of the object points at its first frame, while the background channels and the z-buffer keep the depth of $I_0$.
We generate each window with 30 sampling steps, a classifier-free guidance~\cite{cfg} scale of 3, and flow shift 5, keep the frames at the original timestamps, and concatenate the windows directly into an output of 97 frames at 24 fps.
For example, one benchmark sample with a long projectile arc needs two windows that cover native frames 0 to 70 with 97 frames and 70 to 96 with 29 frames.

\section{Data Pipeline Details}
\label{supp:data}

This section gives the details of \cref{sec:data_pipeline} in the order of the pipeline.

\subsection{Assets and Scenes}
\label{supp:scenes}

\paragraph{Assets.}
The generic objects are 84 Objaverse~\cite{objaverse} meshes whose collision shapes are 22 balls, 35 boxes, and 27 cylinders.
Sketchfab~\cite{sketchfab} supplies 52 models.
Sixteen are household objects such as a crate, a remote control, a can, and a bowl, whose collision shapes combine balls, boxes, and cylinders, and 36 are tables, workbenches, counters, pool tables, and room props, among which the supports collide with their exact mesh surfaces.
Poly Haven~\cite{polyhaven} supplies 95 materials and 509 HDR environment maps.
Each part of a mesh is marked as colliding, as background decoration, as hidden, or as discarded, and the support planes come from the measured surfaces of the support meshes.

\paragraph{Scenes and motions.}
Single objects follow eleven motion types, namely sliding after a push, rolling, wall impact, falling off an edge, dropping onto a support, horizontal and arched projectile flight, sliding down and up a ramp, moving from a ramp onto a floor, and bouncing.
Pairs follow twelve types, namely a moving object hitting or grazing a resting one, frontal and glancing opposed hits, crossing and chasing hits, a falling or thrown ball hitting a supported object, a collision of two airborne balls, and three controls in which the objects never touch, and at least 80\% of the generic pairs interact.
Triples form chained transfers, successive hits, converging hits, and pair controls in which one object stays untouched, on floors, shallow ramps of 8 to 12 degrees, and real tables.
A billiards table, a pinball board with passive pegs, and a marble run add scenes in which balls meet cushions, pegs, tracks, and other balls.

\paragraph{Initial states.}
Placement uses the oriented bounds of the objects, the support normals, and contact clearance, and travel corridors keep incidental scenery out of the motion path.
The initial velocities follow from the intended motion, namely from the sliding distance and duration for a push, the travel time for rolling, the flight time and horizontal extent for a projectile, and the slope, gravity, and friction for a ramp.

\subsection{Simulation, Rendering, and Checks}
\label{supp:simulation}

\paragraph{Simulation and rendering.}
PyBullet~\cite{pybullet} integrates four seconds with a solver frequency that follows the scale and speed of the objects, and the billiards table, pinball board, and marble run keep their own calibrated settings.
Blender~\cite{blender} 3.4 renders with EEVEE at $1280 \times 720$ and 24 fps into 97 frames that include both ends of the four seconds.
Lighting and shadow bias follow the footprint of each object, and exposure adapts to texture measurements while physics, camera, and materials stay fixed.

\paragraph{Physics and event checks.}
The simulator must use exactly the recorded collision shapes, physical parameters, and initial states, a flying object must follow the ballistic path under gravity, mechanical energy must not grow on its own, a resting object must stay in place, and friction forces must stay within the Coulomb limit.
The planned events must happen, so a ball aimed at a wall must hit it, objects must collide in the planned order, a bouncing ball must rebound high enough, and objects planned to pass each other must stay apart.
The camera is chosen from the simulated motion and must show every object large enough and mostly unoccluded in the first frame and at every contact.
A scene that fails any check is replaced by a new sample of the same scene type, and a group is kept only when all of its videos pass.

\paragraph{Record checks.}
Trajectories must be finite, start at time zero with timestamps that increase at the declared frame rate, and keep unit quaternions with continuous signs, and their initial states must match the scene description.
Each object keeps one name across the text prompt, the trajectories, the rendered meshes, and the sweep target.
Videos must match the declared size, frame count, average frame rate, and the timestamp of every frame.
Sweep variants pass the same checks.

\subsection{Sweeps and Splits}
\label{supp:splits}

Each parameter takes five values interpolated within the intersection of its admissible range and a span around the base value, with the base value at the middle.
The inertia tensor scales with the mass.
Each group therefore holds the base video and 12 variants, with two values below and two above the base value for each of mass, sliding friction, and restitution.
For scenes with two or three objects, mass takes 0.25, 0.5, 2, and 4 times the base value.
Splitting by group so that no first frame appears in more than one split gives the counts in \cref{tab:supp_split}.

\begin{table}[t]
  \centering
  \small
  \begin{tabular*}{0.55\linewidth}{@{\extracolsep{\fill}}lrrr@{}}
    \toprule
    \textbf{Objects} & \textbf{Train} & \textbf{Val} & \textbf{Test} \\
    \midrule
    One & 2{,}860 & 158 & 171 \\
    Two & 2{,}783 & 136 & 158 \\
    Three & 1{,}403 & 72 & 63 \\
    \midrule
    All & 7{,}046 & 366 & 392 \\
    \bottomrule
  \end{tabular*}
  \caption{\textbf{Groups in each split.} Each group holds 13 videos.}
  \label{tab:supp_split}
\end{table}

\section{Benchmark and Evaluation Details}
\label{supp:evaluation}

This section gives the details of SCENARIO-Bench in \cref{sec:benchmark}, of the experimental setup in \cref{sec:setup}, of the real images in \cref{sec:real}, and of the ablation in \cref{sec:ablation}.

\subsection{SCENARIO-Bench}
\label{supp:bench}

SCENARIO-Bench takes 20 base scenes each with one, two, and three objects from the test split, and each base scene varies one parameter of its target object.
In all 60 groups, the low value lies below and the high value above the base value, and the three videos of a group share the first frame, the text prompt, and the seed.
The mass groups come from scenes with two or three objects, where mass decides how collisions exchange momentum.
\Cref{tab:supp_bench} lists the groups for each parameter.

\begin{table}[t]
  \centering
  \small
  \begin{tabular*}{0.60\linewidth}{@{\extracolsep{\fill}}lrrrr@{}}
    \toprule
    \textbf{Parameter} & \textbf{One} & \textbf{Two} & \textbf{Three} & \textbf{All} \\
    \midrule
    Mass & 0 & 7 & 15 & 22 \\
    Friction & 9 & 5 & 5 & 19 \\
    Restitution & 11 & 8 & 0 & 19 \\
    \midrule
    All & 20 & 20 & 20 & 60 \\
    \bottomrule
  \end{tabular*}
  \caption{\textbf{Groups of SCENARIO-Bench for each varied parameter and object count.}}
  \label{tab:supp_bench}
\end{table}

\subsection{Metrics}
\label{supp:metrics}

\paragraph{Spatial and temporal alignment.}
The methods generate videos in different formats, as listed in \cref{tab:supp_baselines}.
On SCENARIO-Bench and PhyCo-Sim, we map every output to a common view of $720 \times 415$ pixels at the 49 timestamps of 12 fps within the four seconds and score the 48 frames after the first.
FVD~\cite{fvd} uses the I3D~\cite{i3d} features of the StyleGAN-V~\cite{styleganv} implementation with one clip of 48 frames per video.

\paragraph{Object regions.}
The object region of an object in a frame is the bounding box of its visible surface points projected from the simulated scene, enlarged on every side by 25\% of its longer edge and by at least 12 pixels.
PSNR is computed on this crop, and SSIM~\cite{ssim} and LPIPS~\cite{lpips} are computed after resizing the crop to $256 \times 256$.
Each metric is averaged over the frames of an object, then over the objects of a video, and then over the videos.
On SCENARIO-Bench, the object regions cover 3.8\% of a frame on average.

\paragraph{Contact regions.}
On SCENARIO-Bench, a contact event is a change in the number of contact points that the simulator reports for an object.
On Physion++, it is a change in the number of objects and environment surfaces that touch the object in the collision records.
On PhyCo-Sim, it is a change in the number of scene planes within 2.5\% of the largest bounding-box side of the object, extended by its motion toward the plane in half a frame.
The contact region of an object keeps its object region within 0.25\,s of each of its events, and every method is scored on the same frames.

\paragraph{Sweep score.}
SAM~2~\cite{sam2} propagates the instance masks of the shared first frame through each generated video, and the mask centers enter \cref{eq:sweep_score}.
The ground-truth centers come from object masks rendered from the simulated scene.
Running the same tracker on ground-truth videos gives a 95th percentile error of at most 0.72 pixels, below the floor of one pixel.
A frame in which a generated object vanishes or can no longer be identified counts as a failure, and its squared error takes the larger of the squared denominator of \cref{eq:sweep_score} and the squared ground-truth offset.
Every group has the same weight in the benchmark score.

\subsection{Baselines}
\label{supp:baselines}

\Cref{tab:supp_baselines} lists the output format and sampling settings of every method.
Every baseline runs its released checkpoint, and every method keeps one random seed for all of its videos.
DaS~\cite{das} receives its tracking video rendered from the simulated object points and from background points with valid depth in the first frame.
VACE~\cite{vace} runs once with 16 point trajectories per object and 256 background points and once with depth maps rendered from the simulated scene, and the figures show the depth version on the benchmarks and the trajectory version on real images.
Wan-Move~\cite{wanmove} receives the object trajectories and static background anchors, FlashMotion~\cite{flashmotion} receives the bounding boxes of the simulated objects, and PhyCo~\cite{phyco} receives friction and restitution maps through its released pipeline.
PhysGen~\cite{physgen} receives the first-frame object masks, the physical parameters, and the initial velocities projected into the image in place of the estimates of its perception stage, and its video refinement keeps its default prompt of object names.
Wan2.2 TI2V-5B~\cite{wan} receives only the first frame and the text prompt.

\begin{table}[t]
  \centering
  \small
  \setlength{\tabcolsep}{10pt}
  \begin{tabular}{@{}lll@{}}
    \toprule
    \textbf{Method} & \textbf{Output} & \textbf{Sampling} \\
    \midrule
    Wan2.2 TI2V-5B & $832 \times 480$, 97 frames, 24 fps & 50 steps, guidance 5 \\
    PhyCo & $768 \times 432$, 117 frames, 24 fps & 35 steps, guidance 7 \\
    PhysGen & $1024 \times 576$, 16 frames, 3.75 fps & 50 steps, guidance 8 \\
    FlashMotion & $832 \times 480$, 97 frames, 24 fps & 4 steps \\
    Wan-Move & $832 \times 480$, 81 frames, 20 fps & 40 steps, guidance 5 \\
    DaS & $720 \times 480$, 49 frames, 12 fps & 30 steps, guidance 5 \\
    VACE & $832 \times 480$, 49 frames, 12 fps & 50 steps, guidance 5 \\
    \midrule
    SCENARIO & $832 \times 480$, 97 frames, 24 fps & 30 steps, guidance 3 \\
    \bottomrule
  \end{tabular}
  \caption{\textbf{Output format and sampling settings of every method.} The first 97 frames of PhyCo cover the four seconds of the benchmark. PhysGen places its 16 frames at simulated times from 0 to 4.02\,s, and each timestamp of the benchmark takes the nearest one.}
  \label{tab:supp_baselines}
\end{table}

\subsection{PhyCo-Sim and Physion++}
\label{supp:external}

On these two benchmarks, we compare with the backbone, with DaS and VACE, the strongest baselines on SCENARIO-Bench, and with PhyCo, which introduced PhyCo-Sim.

\paragraph{PhyCo-Sim.}
The rigid-body subset of PhyCo-Sim~\cite{phyco} holds 34 ball drops, 33 collisions between balls and walls, and 33 objects sliding on flat ground, all with a static camera and with gravity, contact, and friction as the only forces after the first frame.
Within each category, we sample equally from the low, middle, and high thirds of the contact coefficient that the category varies.
We build the physical scene of each case from the simulation records released with the dataset.

\paragraph{Physion++.}
The subset of Physion++~\cite{physionpp} takes 100 videos from the official test set, 20 each for collisions under different mass, sliding on platforms under different friction, collisions under different friction, bouncing off walls, and bouncing off platforms.
Every video has a fixed camera and one to three moving rigid bodies, and the clips last between 1 and 2.7 seconds.
We build the physical scene of each case from the meshes, poses, and physical parameters exported with the dataset.
The videos are scored at their native resolution of $256 \times 256$ pixels, and FVD uses the first 12 future frames, which every case provides.

\subsection{Real Images}
\label{supp:real_scene}

We build the geometry and the initial poses of each scene in \cref{sec:real} from its image.
MoGe-2~\cite{moge2} predicts metric depth, and we rescale this depth so that one object of known size in the image matches its true size.
SAM~2~\cite{sam2} segments the objects, the table or furniture top gives the supporting plane, and each object takes a sphere or a box of its known size at the position that the depth gives.
For Physics-IQ~\cite{physicsiq}, we take the switch frame, where the benchmark hands over from the conditioning video to the ground truth, and measure the initial velocities in the frames that precede it.
For Physics 101~\cite{phys101}, we take a frame shortly after the object starts to move and set the initial velocity along its direction of motion in the video.
PhysGen~\cite{physgen} and PhysGen3D~\cite{physgen3d} take the initial force or velocity of an object as an input, and we likewise give the moving object an initial velocity.
For the GAUGE~\cite{gauge} images of \cref{fig:teaser}, the initial velocities come from the metadata of each recording.
We bring every image to our aspect ratio by cropping it or by extending it with FLUX.1 Fill~\cite{flux2024}.

\subsection{Ablated Models}
\label{supp:ablation}

Without physical parameters, the model receives none of them, including the mass, friction, restitution, damping, and inertia of the objects, the friction and restitution of the environment, the mass ratio, and gravity, while the geometry and relative motion of the neighbors remain.
Without local scene relations, the interaction features are computed from the physical parameters alone, without local relation attention.
With neither, the interaction features are removed, and the trajectory features alone condition the backbone.

\section{Additional Results}
\label{supp:results}

This section extends the comparison of \cref{sec:comparison} with the sweep score of every subset of SCENARIO-Bench and an analysis of motion following, and extends \cref{sec:real} with a user study.

\subsection{Sweep Score by Subset}
\label{supp:sweep_subsets}

SCENARIO reaches the highest sweep score in every subset of SCENARIO-Bench.
\Cref{tab:supp_sweep} breaks the sweep score of \cref{tab:bench} down by object count and by the varied parameter.

\begin{table}[t]
  \centering
  \small
  \setlength{\tabcolsep}{8pt}
  \begin{tabular}{@{}lcccccc@{}}
    \toprule
    & \multicolumn{3}{c}{\textbf{Objects}} & \multicolumn{3}{c}{\textbf{Parameter}} \\
    \cmidrule(lr){2-4} \cmidrule(l){5-7}
    \textbf{Method} & \textbf{One} & \textbf{Two} & \textbf{Three} & \textbf{Mass} & \textbf{Fric.} & \textbf{Rest.} \\
    \midrule
    Wan2.2 TI2V-5B & 0.000 & 0.000 & 0.057 & 0.052 & 0.000 & 0.000 \\
    PhyCo & 0.042 & 0.000 & 0.064 & 0.058 & 0.031 & 0.014 \\
    PhysGen & 0.101 & 0.123 & 0.073 & 0.075 & 0.158 & 0.068 \\
    FlashMotion & 0.397 & 0.311 & 0.265 & 0.313 & 0.258 & 0.404 \\
    Wan-Move & 0.418 & 0.540 & 0.494 & 0.470 & 0.605 & 0.379 \\
    DaS & 0.027 & 0.004 & 0.106 & 0.075 & 0.030 & 0.028 \\
    VACE (trajectories) & 0.549 & \underline{0.737} & 0.416 & 0.479 & 0.584 & 0.652 \\
    VACE (depth) & \underline{0.701} & 0.574 & \underline{0.758} & \underline{0.650} & \underline{0.706} & \underline{0.680} \\
    \midrule
    SCENARIO & \textbf{0.903} & \textbf{0.850} & \textbf{0.800} & \textbf{0.812} & \textbf{0.910} & \textbf{0.838} \\
    \bottomrule
  \end{tabular}
  \caption{\textbf{Sweep score on SCENARIO-Bench by object count and by varied parameter.} Bold marks the best result and underline the best baseline.}
  \label{tab:supp_sweep}
\end{table}

\subsection{Motion Following and Offset Size}
\label{supp:motion_following}

\Cref{tab:supp_following} shows that SCENARIO follows the simulated motion most closely and reaches the highest sweep score for both smaller and larger ground-truth offsets.
The center distance is measured between the tracked object centers and the ground-truth centers in the base videos of the groups in which the tracker keeps every object.
The offset size of a group is the denominator of \cref{eq:sweep_score} before the floor, and the table reports the sweep score of every method in two ranges of this size.

\begin{table}[t]
  \centering
  \small
  \setlength{\tabcolsep}{10pt}
  \begin{tabular}{@{}lccc@{}}
    \toprule
    \multirow{2}{*}[-2.5pt]{\textbf{Method}} & \textbf{Center} & \multicolumn{2}{c}{$\boldsymbol{S}$ \textbf{by offset (px)} $\boldsymbol{\uparrow}$} \\
    \cmidrule(l){3-4}
    & \textbf{distance (px)} $\boldsymbol{\downarrow}$ & \textbf{5 to 60} & \textbf{above 60} \\
    \midrule
    Wan2.2 TI2V-5B & 122.1 & 0.000 & 0.000 \\
    PhyCo & 125.4 & 0.007 & 0.028 \\
    PhysGen & 83.3 & 0.069 & 0.165 \\
    FlashMotion & 51.3 & 0.293 & 0.379 \\
    Wan-Move & 29.5 & 0.411 & 0.677 \\
    DaS & 78.3 & 0.029 & 0.043 \\
    VACE (trajectories) & 11.1 & 0.523 & \underline{0.690} \\
    VACE (depth) & \underline{8.2} & \underline{0.707} & 0.677 \\
    \midrule
    SCENARIO & \textbf{2.4} & \textbf{0.800} & \textbf{0.957} \\
    \bottomrule
  \end{tabular}
  \caption{\textbf{Motion following and sweep score by offset size on SCENARIO-Bench.} The center distance is averaged over the base videos, and the sweep score $S$ covers 36 and 21 groups with smaller and larger ground-truth offsets. The other 3 groups have ground-truth offsets below 5 pixels. Bold marks the best result and underline the best baseline.}
  \label{tab:supp_following}
\end{table}

\subsection{User Study on Real Images}
\label{supp:user_study}

\begin{table}[t]
  \centering
  \small
  \begin{tabular*}{0.65\linewidth}{@{\extracolsep{\fill}}lccc@{}}
    \toprule
    & \multicolumn{2}{c}{\textbf{Part A}} & \textbf{Part B} \\
    \cmidrule(lr){2-3} \cmidrule(l){4-4}
    \textbf{Method} & \textbf{Setting} & \textbf{Quality} & \textbf{Accuracy} \\
    \midrule
    DaS & 94.4 & 91.7 & 61.1 \\
    VACE (trajectories) & 91.7 & 93.8 & 77.8 \\
    PhyCo & 96.5 & 93.1 & 50.0 \\
    \midrule
    SCENARIO & & & 94.4 \\
    \bottomrule
  \end{tabular*}
  \caption{\textbf{User study on real images.} The Part A columns give the percentage of judgments that prefer SCENARIO over each method on how well the motion fits the stated physical setting and on visual quality. The Part B column gives the percentage of judgments that identify the variant with the heavier, bouncier, or more slippery object from the videos of each method, where chance is 50\%.}
  \label{tab:supp_user_study}
\end{table}

Participants prefer SCENARIO over every baseline in 91.7\% to 96.5\% of the judgments on the stated physical setting and in 91.7\% to 93.8\% of the judgments on visual quality, as \cref{tab:supp_user_study} shows.
From the videos of SCENARIO, they identify which variant has the heavier, bouncier, or more slippery object in 94.4\% of the judgments, whereas from the videos of the baselines they do so in at most 77.8\%.

We conduct this study with 24 participants who are not involved in this work, on all 36 sweeps of \cref{sec:real}, and compare SCENARIO with DaS, VACE, and PhyCo.
Every clip shows the first 4~s at $832 \times 480$ and 12 fps without method names, and the two clips of a pair play side by side in random order.

In Part A, each pair shows SCENARIO and one baseline on the same variant with a sentence that states the physical setting, and participants choose the video whose motion fits the setting better and the video with the more stable and realistic objects and scene.
In Part B, each pair shows the low and the high variant of one sweep from the same method, and participants choose the video with the heavier, bouncier, or more slippery object.
We split the 216 pairs of Part A and the 144 pairs of Part B into 12 forms with two attention checks each, every participant answers one form, and all 24 participants pass both checks, so every entry of Part A counts 144 judgments and every entry of Part B counts 72 judgments.

\section{Failure Cases}
\label{supp:failures}

\begin{figure*}[!t]
  \centering
  \includegraphics[width=\textwidth]{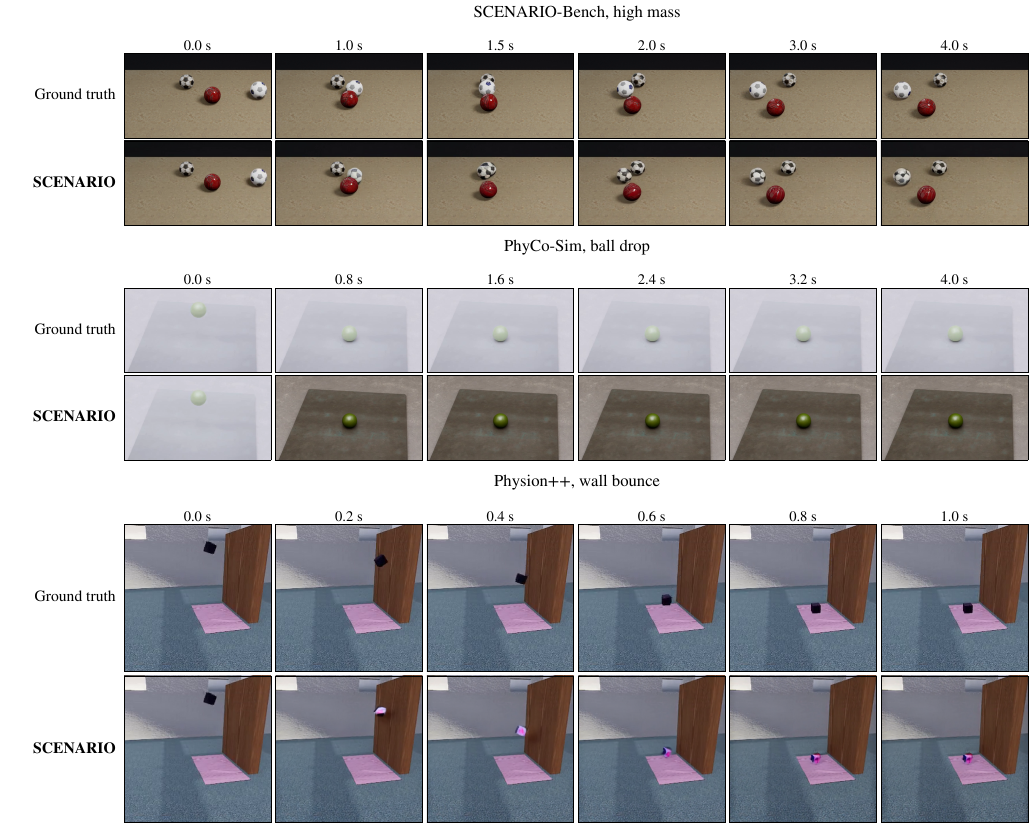}
  \caption{\textbf{Failure cases of SCENARIO.} One case from each benchmark, with the ground truth above SCENARIO. One soccer ball takes on the pattern of the other on SCENARIO-Bench, the whole scene darkens on PhyCo-Sim, and the falling cube turns pink on Physion++.}
  \label{fig:supp_failures}
\end{figure*}

\Cref{fig:supp_failures} shows one failure case from each benchmark, and each paragraph below gives its likely cause and a possible remedy.

\paragraph{SCENARIO-Bench.}
In the high-mass variant of a group with three balls, SCENARIO moves all three balls along their ground-truth paths, but after the two soccer balls touch, the ball that rolls to the left takes on the black pattern of the other one.
While they touch, both balls fall into the same few latent tokens, where the backbone can mix their appearance.
Conditioning each object on an appearance feature taken from its region of the first frame could keep the two patterns apart.

\paragraph{PhyCo-Sim.}
In a ball drop, SCENARIO darkens the whole scene after the first frame and renders the pale ball dark green, although the ball follows its ground-truth path.
This first frame is nearly white and overexposed, whereas our renders adapt their exposure to the scene textures, so the training data holds few such frames.
Training with random changes of exposure and white balance could close this gap.

\paragraph{Physion++.}
In a wall bounce, the small dark cube turns pink when it hits the wall and keeps a pink tint until it rests on the pink mat.
The cube covers only a few latent tokens and the impact falls within one latent frame, so the conditions describe this contact only coarsely.
Generating small objects at a higher resolution, for example in a cropped window around each contact, or subdividing the timeline more finely near impacts could reduce such failures.

\section{Limitations and Future Work}
\label{supp:limitations}

SCENARIO takes the physical scene, namely the collision geometry, the physical parameters, and the simulated trajectories, as input, and all three benchmarks provide this scene from simulation.
The conditions rely on the depth of the first frame under a fixed camera, and rendering the static scene along a camera trajectory would extend them to moving cameras.
Deformable and articulated objects and fluids require simulators beyond rigid-body dynamics, and the trajectory-anchored relations carry over to them because the anchors only need surface points and their velocities, which such simulators also provide.

\section{Additional Qualitative Results}
\label{supp:visual}

This section compares SCENARIO with the baselines on all three benchmarks and on real images.

\begin{figure*}[!t]
  \centering
  \includegraphics[width=\textwidth]{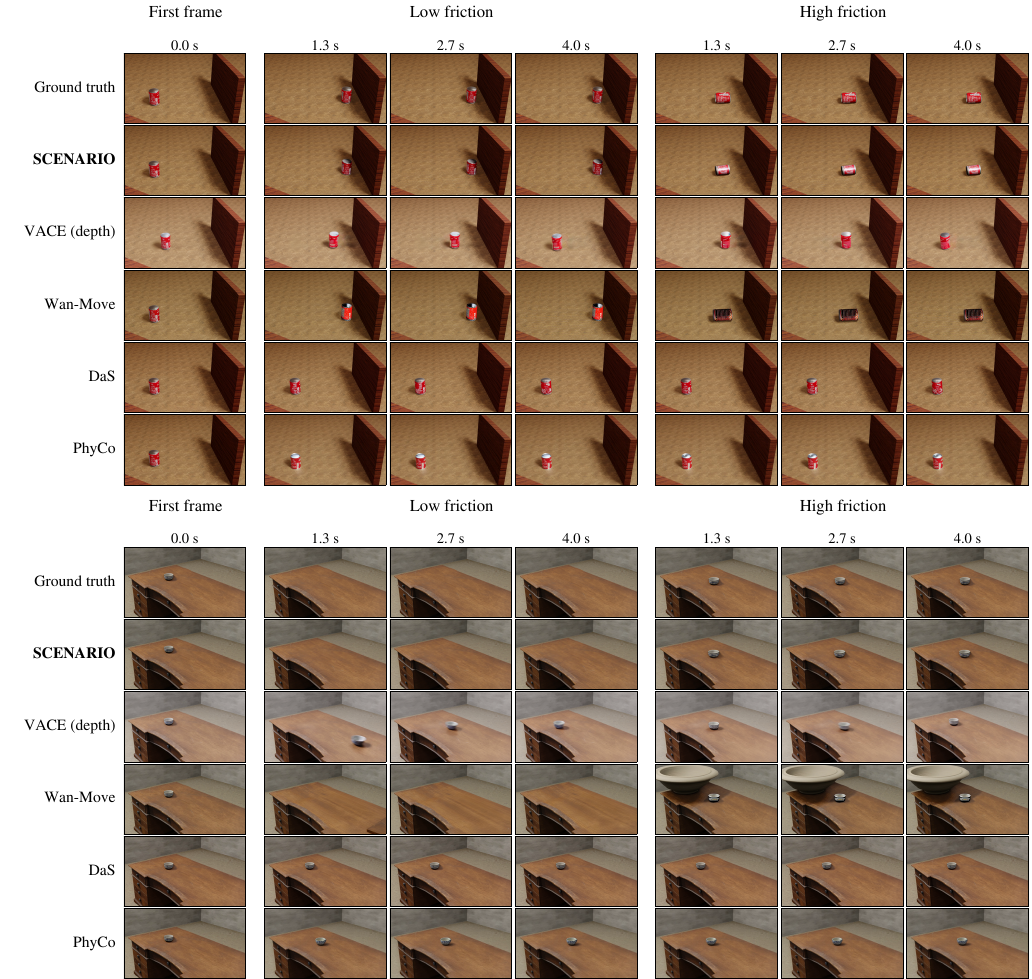}
  \caption{\textbf{Qualitative comparison on two friction groups of SCENARIO-Bench.} With low friction the can slides to the wall and the bowl slides off the table, whereas with high friction the can tips over on the way and the bowl stays on the table.}
  \label{fig:supp_friction}
\end{figure*}

\begin{figure*}[!t]
  \centering
  \includegraphics[width=\textwidth]{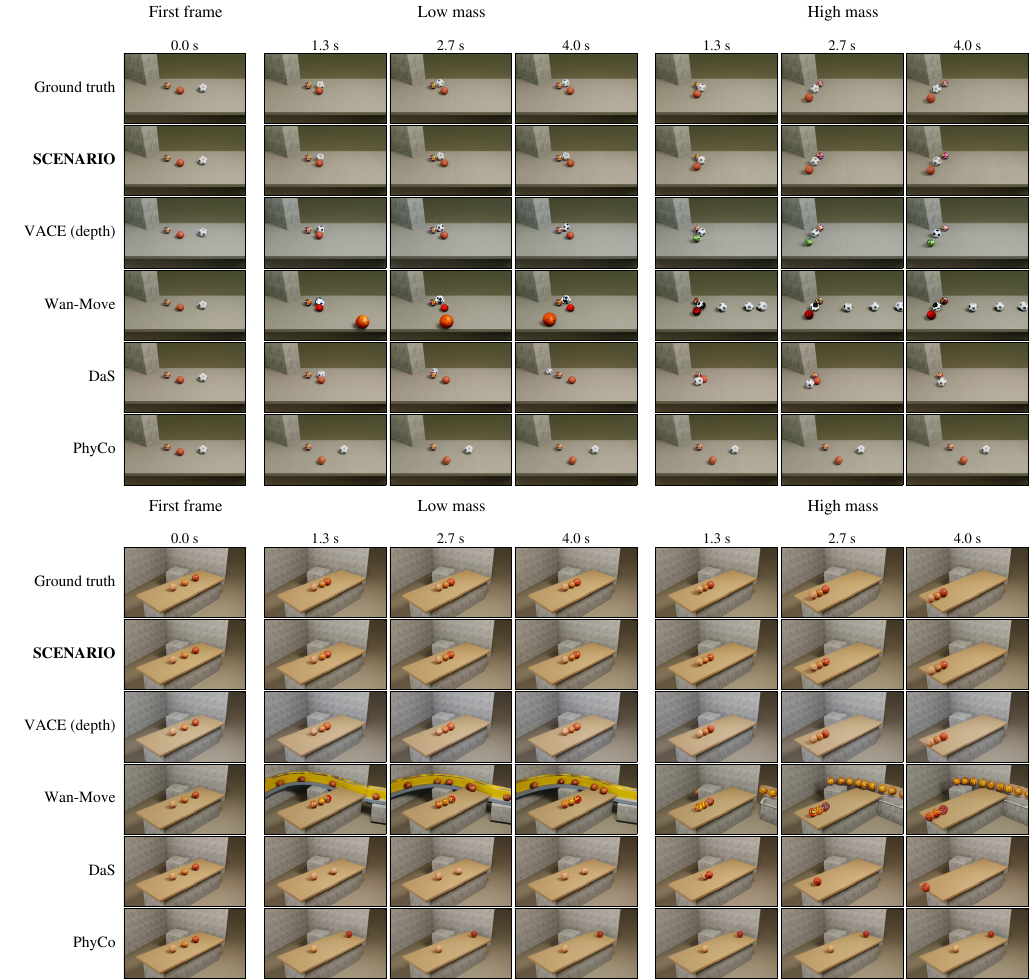}
  \caption{\textbf{Qualitative comparison on two mass groups of SCENARIO-Bench.} In both groups, the rolling ball stops at the two resting balls with low mass but drives them away with high mass.}
  \label{fig:supp_mass}
\end{figure*}

\begin{figure*}[!t]
  \centering
  \includegraphics[width=\textwidth]{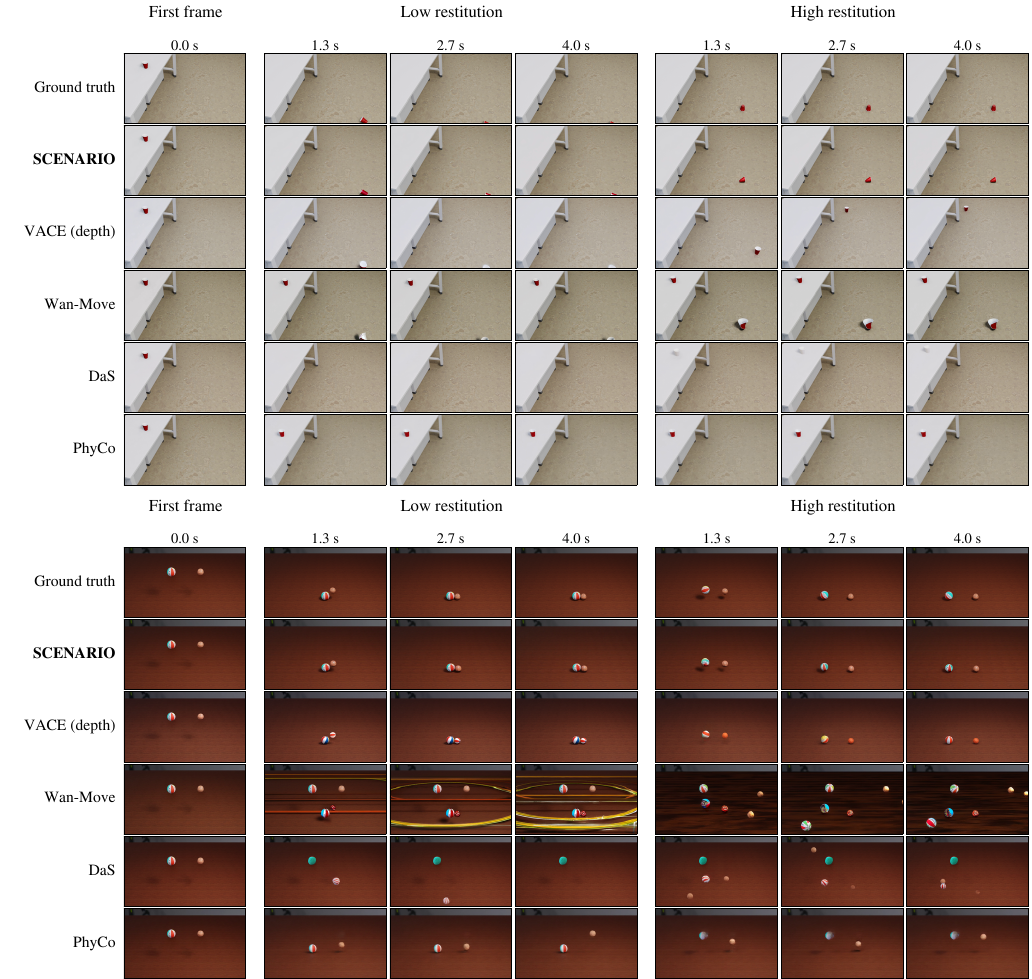}
  \caption{\textbf{Qualitative comparison on two restitution groups of SCENARIO-Bench.} The cup that falls off the table comes to rest at different places on the floor, and the two balls that collide in the air stay together with low restitution but separate with high restitution.}
  \label{fig:supp_restitution}
\end{figure*}

\paragraph{SCENARIO-Bench.}
In the six groups of \cref{fig:supp_friction,fig:supp_mass,fig:supp_restitution}, SCENARIO reproduces both outcomes of every group and keeps the appearance of the objects.
The groups cover friction, mass, and restitution with one to three objects.
The baselines miss outcomes or alter the objects, as in \cref{fig:qualitative}.

\begin{figure*}[!t]
  \centering
  \includegraphics[width=\textwidth]{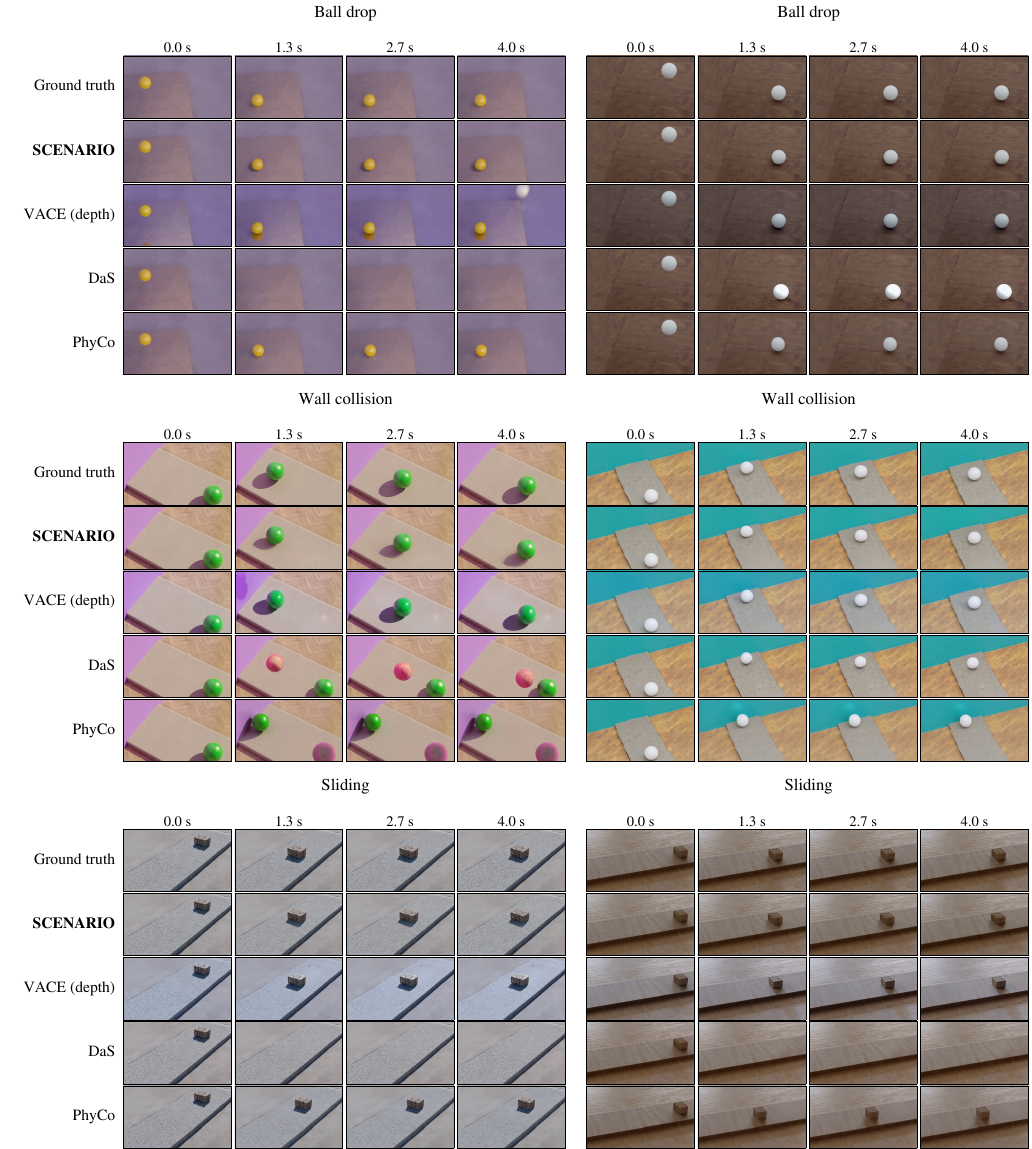}
  \caption{\textbf{Qualitative comparison on PhyCo-Sim.} Two ball drops, two wall collisions, and two sliding cases.}
  \label{fig:supp_phycosim}
\end{figure*}

\paragraph{PhyCo-Sim.}
In the six cases of \cref{fig:supp_phycosim}, SCENARIO follows the ground-truth motion of every object.
The baselines shift the scene colors, lose objects, or misplace them.

\begin{figure*}[!t]
  \centering
  \includegraphics[width=\textwidth]{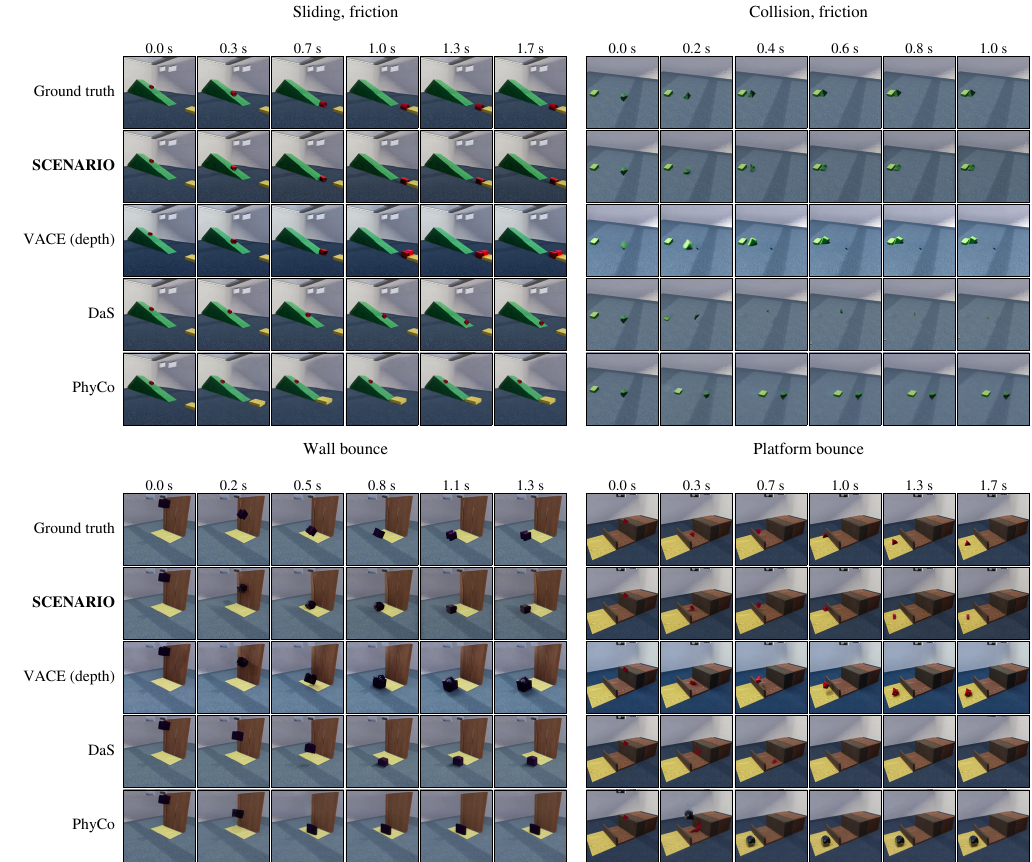}
  \caption{\textbf{Qualitative comparison on Physion++.} A slide down a ramp, a sliding collision, a wall bounce, and a platform bounce.}
  \label{fig:supp_physion}
\end{figure*}

\paragraph{Physion++.}
In the four cases of \cref{fig:supp_physion}, SCENARIO follows the ground-truth motion of every object.
The baselines alter or lose the objects, or depart from the ground-truth motion.

\begin{figure*}[!t]
  \centering
  \includegraphics[width=\textwidth]{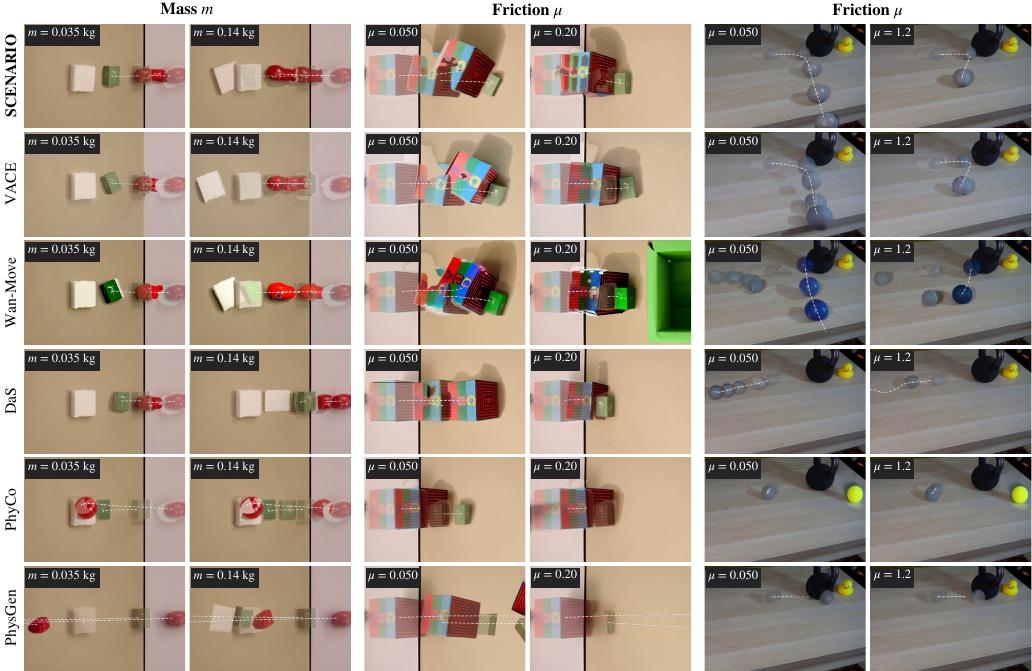}
  \caption{\textbf{Qualitative comparison on real images, first part.} On the left, a toy slides into a green box with low and high mass. In the middle, a cardboard cube slides into the same green box with low and high friction. On the right, a gray ball rolls into a kettlebell with low and high friction. The left and middle images come from Physics 101 and the right one from Physics-IQ. Each panel overlays the frames of one video at the same instants, and dashed lines trace the objects tracked in that video.}
  \label{fig:supp_real_a}
\end{figure*}

\begin{figure*}[!t]
  \centering
  \includegraphics[width=\textwidth]{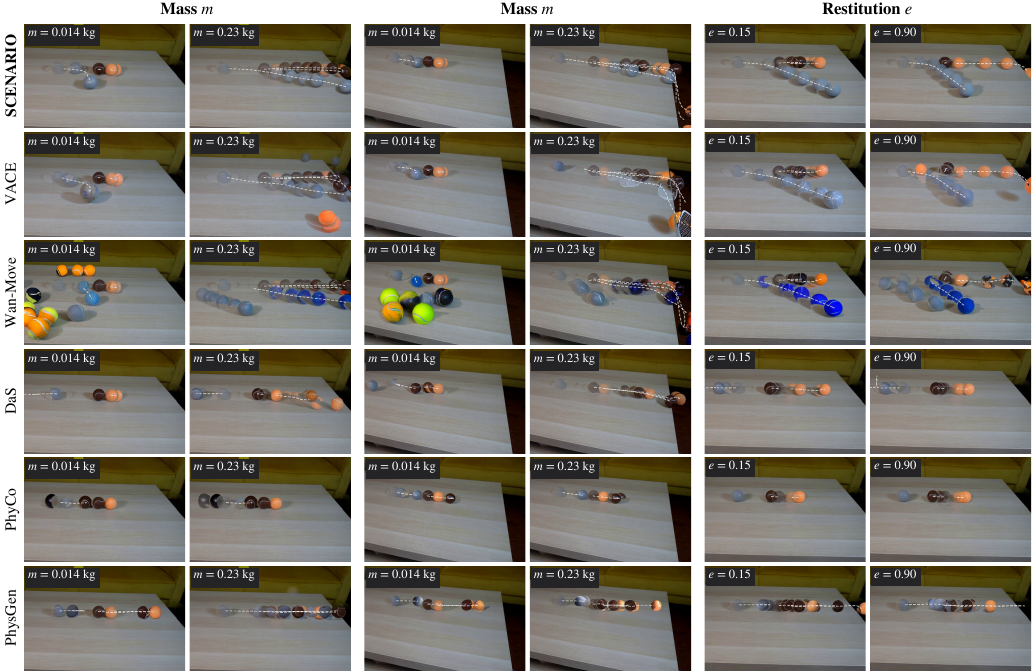}
  \caption{\textbf{Qualitative comparison on real images, second part.} On the left and in the middle, a gray ball strikes two balls with low and high mass. On the right, a gray ball strikes two balls with low and high restitution. All three images come from Physics-IQ. Each panel overlays the frames of one video at the same instants, and dashed lines trace the objects tracked in that video.}
  \label{fig:supp_real_b}
\end{figure*}

\paragraph{Real images.}
In the six sweeps of \cref{fig:supp_real_a,fig:supp_real_b}, SCENARIO keeps the appearance of the objects and produces a distinct outcome for each value.
The baselines change object colors, add extra objects, or produce the same motion for both values.


\end{document}